\documentclass{article}

\usepackage[accepted]{icml2026}

\usepackage[utf8]{inputenc} 
\usepackage[T1]{fontenc}    
\usepackage{microtype}      
\usepackage{xspace}         

\usepackage[table,xcdraw]{xcolor}
\usepackage{colortbl}

\usepackage{icml2026}

\usepackage{amsmath}
\usepackage{amssymb}
\usepackage{mathtools}
\usepackage{amsthm}
\usepackage{amsfonts}
\usepackage{nicefrac}       

\usepackage{booktabs}      
\usepackage{multirow}
\usepackage{array}
\usepackage{tabularx}
\usepackage{tablefootnote}  

\usepackage{graphicx}
\usepackage{subcaption}     
\usepackage[font=small,labelfont=bf]{caption} 
\usepackage[inkscape]{svg} 

\usepackage{listings}
\usepackage{mdframed}

\usepackage{enumerate}
\usepackage{enumitem}
\usepackage{chngcntr}

\usepackage{url}
\usepackage{hyperref}
\usepackage[capitalize,noabbrev]{cleveref} 

\usepackage{cuted}
\usepackage{capt-of} 

\usepackage{tabularx}
\newcolumntype{Y}{>{\centering\arraybackslash}X} 

\theoremstyle{plain}

\theoremstyle{definition}

\theoremstyle{remark}

\usepackage[textsize=tiny]{todonotes}

\icmltitlerunning{Forget-It-All}

\begin{document}

\twocolumn[
  \icmltitle{Forget-It-All: Multi-Concept Machine Unlearning via Concept-Aware Neuron Masking}

  \begin{icmlauthorlist}
    \icmlauthor{Kaiyuan Deng}{ua}
    \icmlauthor{Bo Hui}{tulsa}
    \icmlauthor{Gen Li}{clemson}
    \icmlauthor{Jie Ji}{clemson}
    \icmlauthor{Minghai Qin}{wdc}
    \icmlauthor{Geng Yuan}{uga}
    \icmlauthor{Xiaolong Ma}{ua}
  \end{icmlauthorlist}

  \icmlaffiliation{ua}{The University of Arizona}
  \icmlaffiliation{tulsa}{The University of Tulsa}
  \icmlaffiliation{clemson}{Clemson University}
  \icmlaffiliation{wdc}{Western Digital Corporation}
  \icmlaffiliation{uga}{University of Georgia}

  \icmlcorrespondingauthor{Kaiyuan Deng}{kaiyuan0415@arizona.edu}
]

\printAffiliationsAndNotice{}  


\begin{strip}
    \centering
    \vspace{-4.4em} 
    \includegraphics[width=\textwidth]{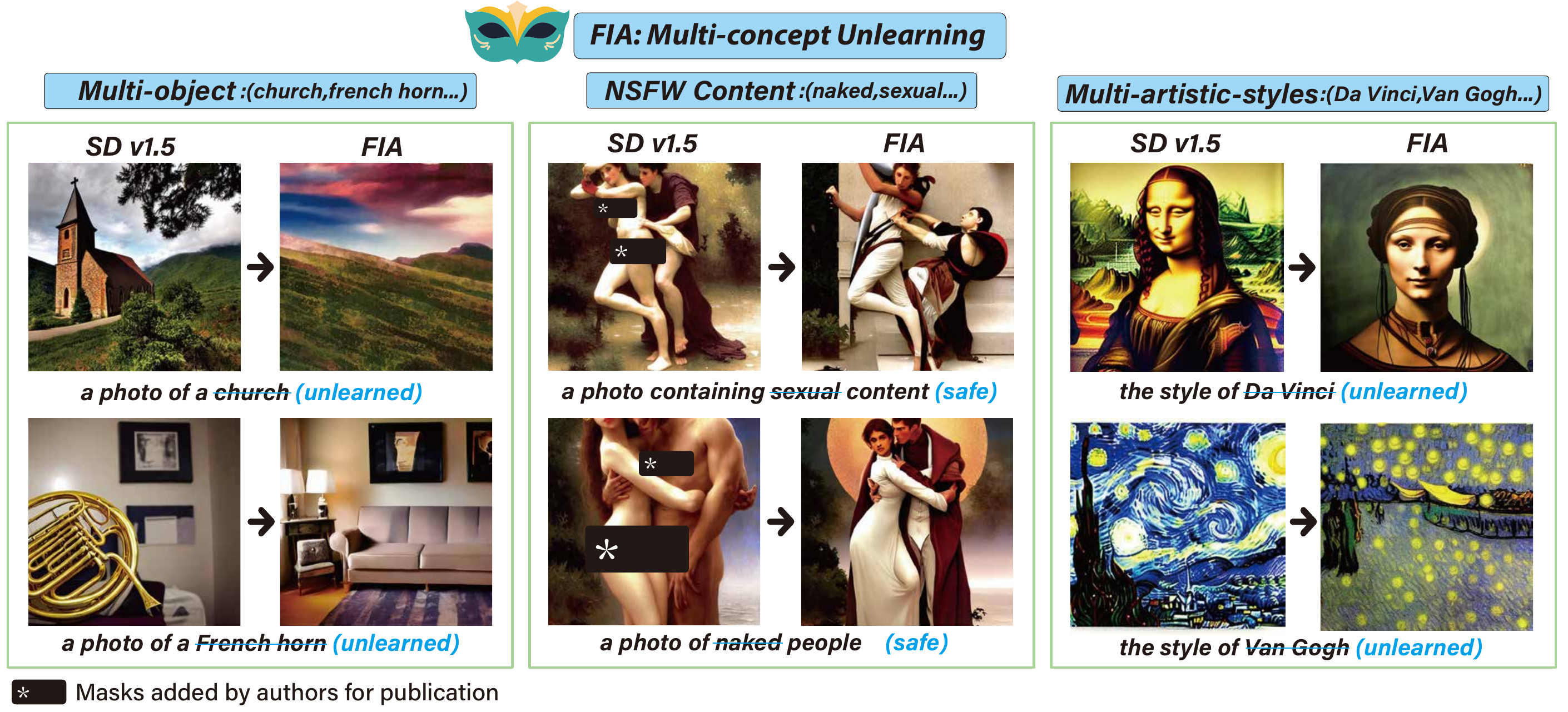}
    \captionof{figure}{
  The proposed FIA framework enables simultaneous multi-concept unlearning in text-to-image models. We demonstrate the unlearning effects of 2 concepts in the multi-concept scenario with FIA (more comprehensive results are shown in Section~\ref{exp}). It shows that FIA can (i) unlearn multiple undesired objects, (ii) prevent the generation of explicit content, and (iii) mitigate artwork copyright issues. This figure illustrates that FIA not only achieves robust multi-concept unlearning but also preserves the generation quality.
    }
    \label{fig:c1}
\end{strip}

\begin{abstract}
The widespread adoption of text to image (T2I) diffusion models has raised concerns about their potential to generate copyrighted, inappropriate, or sensitive imagery. As a practical solution, \emph{machine unlearning} aims to erase unwanted concepts without retraining from scratch. While most existing methods are effective for single concept unlearning, they often struggle when removing multiple concepts, causing significant challenges in unlearning effectiveness, generation quality, and sensitivity to hyperparameters and datasets. We take a unique perspective on multi-concept unlearning by leveraging model sparsity and propose the \underline{F}orget \underline{I}t \underline{A}ll (FIA) framework. FIA first introduces \emph{Contrastive Concept Saliency} to quantify each weight connection's contribution to a target concept. It then identifies \emph{Concept Sensitive Neurons} by combining temporal and spatial information, ensuring that only neurons consistently responsive to the target concept are selected. Finally, FIA constructs masks from the identified neurons and fuses them into a unified multi concept mask, where \emph{Concept Agnostic Neurons} that broadly support general content generation are preserved while concept specific neurons are pruned to remove the targets. FIA is training free and requires minimal hyperparameter tuning for new tasks, enabling plug and play use. Extensive experiments across three distinct unlearning tasks demonstrate that FIA achieves more reliable multi-concept unlearning, improving forgetting effectiveness while maintaining generation fidelity and quality. Code is available at \url{https://github.com/kaiyuan02415/Forget-It-All}
\end{abstract}

\section{Introduction}
\label{intro}

Text-to-image (T2I) diffusion models~\citep{rombach2022high, ma2024sit, podell2023sdxl, saharia2022photorealistic, nichol2021glide, ding2022cogview2, zhou2022towards} have achieved strong performance and versatility across diverse real-world applications~\citep{kazerouni2022diffusion, xing2024survey}. However, these models also pose significant ethical and legal risks~\citep{weidinger2021ethical, brundage2018malicious, tolosana2020deepfakes, solaiman2021process, xu2024machine, marino2025bridge, liu2024machine}, including copyright infringement and the generation of harmful content, underscoring the need for practical mitigation strategies. A promising solution is machine unlearning~\citep{bourtoule2021machine, guo2019certified, graves2021amnesiac, jang2022knowledge, nguyen2022survey, zhang2023review, xiao2025right}, which selectively removes the influence of problematic data from pre-trained models without degrading overall performance. By preventing the generation of specified concepts, machine unlearning enhances copyright compliance and model safety while avoiding costly retraining from scratch.

Most existing machine unlearning (MU) methods~\citep{zhang2024forget, fan2023salun, heng2023selective, kumari2023ablating, gandikota2023erasing, lyu2024one, lu2024mace, zhao2024separable, xiao2025efficient} are primarily designed to erase single concepts. Fine-tuning-based methods~\citep{zhang2024forget, fan2023salun, kumari2023ablating, gandikota2023erasing} achieve concept erasure by updating the cross-attention layers of diffusion models. In contrast, training-free approaches, such as weight editing~\citep{chavhan2024conceptprune} and Elastic Weight Consolidation~\citep{heng2023selective}, remove concepts without fine-tuning. However, these single-concept unlearning methods struggle in real-world scenarios involving multiple concepts. When applied sequentially, they often cause the model to re-acquire forgotten concepts or suffer degraded generative quality.

To achieve multi-concept unlearning, SPM~\citep{lyu2024one} and MACE~\citep{lu2024mace} employ low-capacity adapters or LoRA weight merging to erase multiple concepts. Meanwhile, ESD~\citep{gandikota2023erasing} removes concept combinations via an LLM-derived concept graph and adversarial feature decoupling, while SepME~\citep{zhao2024separable} erases concepts through residual extraction and cross-attention nullspace decomposition. Furthermore, UCE~\citep{gandikota2024unified} performs a closed-form edit of cross-attention weights to remove concepts.
Despite this progress, existing multi-concept unlearning methods still face two major challenges:
\emph{(i)} they either degrade generative performance after unlearning or fail to fully remove all target concepts, limiting the balance between unlearning efficacy and generation quality;
\emph{(ii)} most methods rely on fine-tuning and are thus sensitive to hyperparameters, leading to complex tuning, increased computational cost, and a risk of overfitting.

In this paper, we propose \underline{\textbf F}orget-\underline{\textbf I}t-\underline{\textbf A}ll (\textbf{FIA}), a training free framework that can simultaneously forget \emph{arbitrary sets of concepts} while retaining the model’s generative quality (as shown in Figure~\ref{fig:c1}). 
%
FIA introduces \emph{Contrastive Concept Saliency} to quantify each weight connection’s contribution to a target concept. It identifies \emph{Concept-Sensitive Neurons} by aggregating neuron responses across denoising timesteps and ranking them channel-wise and cross-channel, ensuring reliable association with the target concept. Finally, FIA constructs per-concept masks and applies a multi-concept fusion strategy that preserves \emph{Concept-Agnostic Neurons}, which respond broadly across concepts to maintain generative quality, while pruning concept-specific neurons.
FIA requires neither fine-tuning nor concept mapping, and operates with only a small set of easily controlled parameters. Its plug-and-play design makes it rapidly deployable across diverse unlearning tasks.
We conducted comprehensive experiments on various datasets
to evaluate FIA’s effectiveness in three distinct unlearning scenarios: (1) \emph{multi‑object unlearning}; (2) \emph{multi‑artist-style unlearning}; (3) \emph{explicit content unlearning}. Our experimental results demonstrate that FIA consistently outperforms state‑of‑the‑art approaches across a range of unlearning tasks. 
The key contributions of our work are threefold:

\begin{itemize}[leftmargin=1.5em, itemsep=1pt, topsep=1.5pt] 
\item We present a novel perspective that connects model sparsity with multi-concept unlearning, where each concept corresponds to a distinct neuron mask. To the best of our knowledge, this is the first work to explore multi-concept unlearning through unstructured neuron masking, enabling fine-grained control over concept forgetting.
\item 
The proposed FIA multi-concept unlearning framework is training-free and works in a plug-and-play paradigm. It uncovers and exploits two distinct types of neurons: concept-sensitive neurons, which are selectively pruned to remove target concepts, and concept-agnostic neurons, which are preserved to maintain generative quality. 
\item 
We demonstrate that FIA generalizes across tasks. 
We conduct extensive experiments on multiple unlearning tasks, 
showing FIA achieves state-of-the-art unlearning performance at under 0.3\% overall sparsity. 
Such robust performance paves the way for more regulated T2I applications.
\end{itemize}

\section{Related Work}
\label{related}
\textbf{Machine Unlearning} (MU)~\citep{bourtoule2021machine, guo2019certified, graves2021amnesiac, jang2022knowledge, podell2023sdxl, vatter2023evolution} has emerged to address removal requests and data privacy concerns without requiring costly full retraining. Its goal is to erase the influence of specific data or learned concepts while preserving the model’s overall performance. In diffusion models, MU approaches fall into two categories: single-concept unlearning and multi-concept unlearning.

\textbf{Single-concept Unlearning.}  
Finetuning-based approaches have been widely explored. FMN~\citep{zhang2024forget} leverages attention re-steering for concept forgetting, while SalUn~\citep{fan2023salun} modifies critical weights through gradient-based saliency. AC~\citep{kumari2023ablating} aligns generated images for a target concept with a broader anchor concept before ablating, and SA~\citep{heng2023selective} adopts continual learning principles for concept erasure. ESD~\citep{gandikota2023erasing} fine-tunes the model using negative guidance to steer outputs away from undesired content, while MS~\citep{jia2023model} utilizes sparse training to unlearn concepts. However, all these methods are sensitive to hyperparameters and the characteristics of the training data.
In contrast, training-free methods have also been proposed. SLD~\citep{schramowski2023safe} injects safety guidance into latent space to suppress unwanted outputs, but latent biases can persist since weights are unchanged. ConceptPrune~\citep{chavhan2024conceptprune} removes neurons tied to an undesired concept; however, handling multiple concepts is difficult due to complex neuron interactions.

\textbf{Multi-concept Unlearning.} 
MACE~\citep{lu2024mace} uses closed-form cross-attention refinement with multiple LoRA modules, while SPM~\citep{lyu2024one} offers a non-invasive erasure strategy. COGFD~\citep{gandikota2023erasing} leverages LLM-generated concept graphs for harmful concept elimination, SepME~\citep{zhao2024separable} employs weight decoupling for targeted erasure, SCULPT~\citep{li2025sculpting} introduces dynamic masks with concept-aware optimization, and Stereo~\citep{srivatsan2025stereo} proposes a two-stage adversarially robust erasure framework. However, as the number of concepts grows, fine-tuning overhead increases, hyperparameter tuning becomes more complex, and balancing unlearning effectiveness with generative quality becomes increasingly challenging.
In contrast, training-free methods such as UCE~\citep{gandikota2024unified}, ScaPre~\citep{deng2026forget}, and SPEED~\citep{li2025speed} introduce closed-form solutions to modify cross-attention weights for efficient concept erasure. However, they depend on precomputed concept embeddings to guide the edits, and direct parameter manipulation can inadvertently degrade image quality.

\begin{figure*}[t]
  \centering
  \includegraphics[width=1\textwidth]{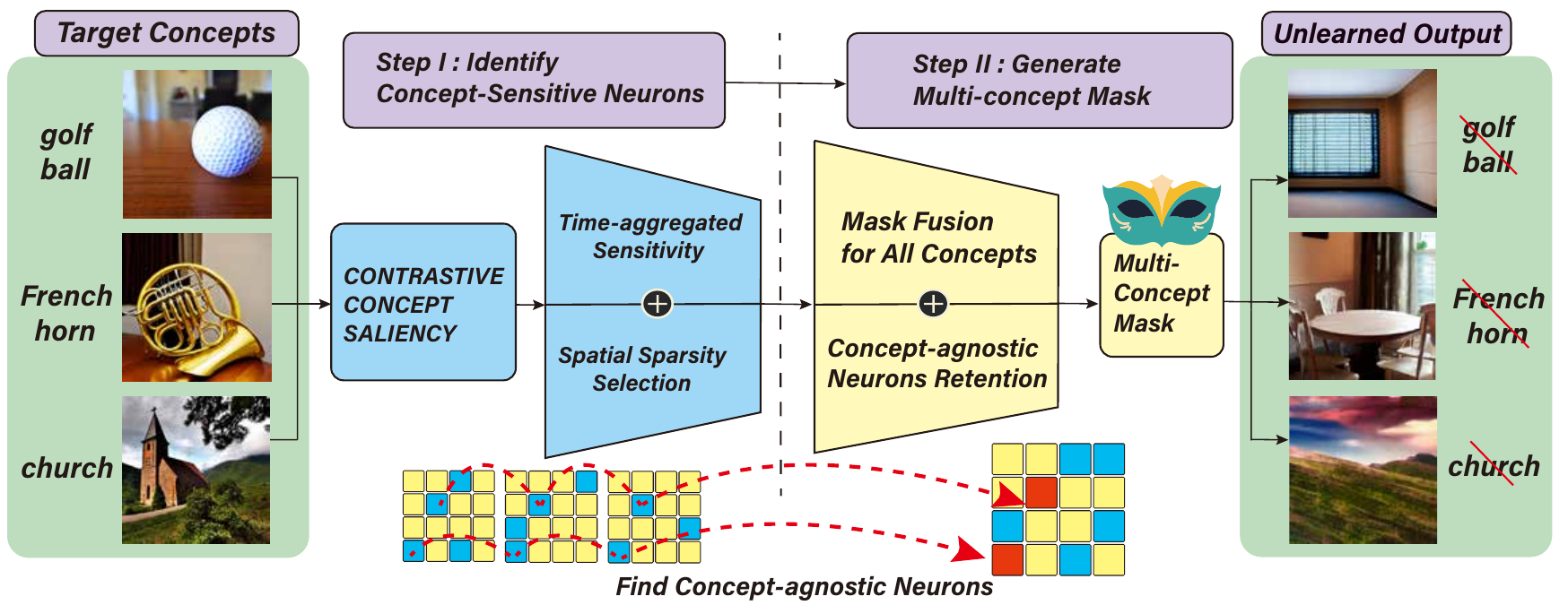}
  \caption{
Overview of our unlearning framework (illustrated with golf ball, French horn, and church). 
We first compute \emph{Contrastive Concept Saliency} to quantify neuron responses to target concepts. 
These scores are aggregated over time and refined with spatial sparsity to identify \emph{Concept-Sensitive Neurons}. 
Finally, we generate per-concept masks and fuse them into a multi-concept mask while preserving \emph{concept-agnostic neurons}.
  }
  \label{fig:m1}
  \vspace{-1em}
\end{figure*}

\section{Method}
\label{method}

In this section, we present FIA, a training-free framework for multi-concept unlearning.  
We begin in Section~\ref{m:sali} by formulating a unified energy-based saliency and introducing \emph{Contrastive Concept Saliency}, which quantifies each weight connection’s contribution to a target concept.  
Then, in Section~\ref{m:iden}, we describe how to identify \emph{Concept-Sensitive Neurons} by integrating temporal sparsity and spatial sparsity, ensuring that only neurons truly responsive to the target concept are retained.  
Finally, in Section~\ref{m:fuse}, we construct per-concept masks from the identified neurons, introduce the notion of \emph{Concept-Agnostic Neurons}, and design a fusion strategy that combines all single-concept masks into a unified multi-concept mask.  
This strategy explicitly preserves concept-agnostic neurons to maintain core generative capacity, while pruning only those neurons that are truly specific to the target concepts.
An overview of the entire FIA pipeline is illustrated in Figure~\ref{fig:m1}.

\subsection{Contrastive Concept Saliency}
\label{m:sali}

To comprehensively quantify each weight neuron's contribution to concept generation, we propose a unified energy saliency that simultaneously accounts for the structural capacity of weights, the activation magnitude of input features, and the effectiveness of signal transmission at each denoising step. 
Consider a linear layer $\ell$ with weight matrix $W\!\in\!\mathbb{R}^{C_{\mathrm{out}}\times C_{\mathrm{in}}}$, where $C_{\mathrm{in}}$ and $C_{\mathrm{out}}$ are the numbers of input and output channels. 
At timestep $t$, the activations are denoted as $X_{\ell,t}\!\in\!\mathbb{R}^{N\times C_{\mathrm{in}}}$, where $N$ is the total number of positions after flattening the batch and spatial or token dimensions. 
We let $X_{\ell,t,j}\!\in\!\mathbb{R}^{N}$ represent the activation vector of the $j$-th input channel, and the response of the $i$-th output channel is 
$Y_{\ell,t,i} = \sum_{j'=1}^{C_{\mathrm{in}}} W_{\ell,i,j'} X_{\ell,t,j'}$. 
For the weight connection between input channel $j$ and output channel $i$, the saliency score at step $t$ is defined as:
\begin{equation}
\label{eq:unified_energy}
U_{\ell,t,i,j}
= |W_{\ell,i,j}| \cdot \|X_{\ell,t,j}\|_2 \cdot 
\frac{|\langle X_{\ell,t,j}, Y_{\ell,t,i} \rangle|}
     {\|X_{\ell,t,j}\|_2 \cdot \|Y_{\ell,t,i}\|_2 + \varepsilon},
\end{equation}
where $\|X_{\ell,t,j}\|_2$ and $\|Y_{\ell,t,i}\|_2$ are the $\ell_2$-norms of the input and output activations, $\langle \cdot,\cdot \rangle$ denotes the Euclidean inner product, and $\varepsilon$ is a small constant to avoid numerical instability. 

From a theoretical perspective, this metric serves as a training-free approximation of neuron importance based on \emph{effective signal propagation}~\cite{lee2018snip, wang2020picking, xiao2025lightcache, li2024neurrev}.  While gradient-based pruning methods typically rely on the loss-gradient-weight product, our formulation approximates the contribution to the latent representation without requiring backpropagation.  Specifically, $|W_{\ell,i,j}| \cdot \|X_{\ell,t,j}\|_2$ measures the \emph{magnitude of signal flow}, prioritizing connections combining high structural capacity with strong input activations.  Crucially, the cosine similarity term penalizes neurons that activate strongly but transmit orthogonal (noise-like) signals not constructively contributing to output features.  By integrating these aspects, $U_{\ell,t,i,j}$ effectively identifies neurons both structurally significant and functionally coherent for the target concept.

Building on this unified energy formulation, we further compute a saliency score to identify neurons that respond specifically to a target concept rather than to general background patterns. 
For each target concept, we evaluate $U_{\ell,t,i,j}$ under two types of textual prompts:  
\begin{itemize}[nosep,leftmargin=*,label={--}]
  \item \emph{Concept prompt}: explicitly contains the target concept  
        (e.g., ``a golf ball on the table'').  
  \item \emph{Base prompt}: describes only the surrounding context  
        (e.g., ``a table'').  
\end{itemize}
Based on these responses, we introduce \textbf{\emph{Contrastive Concept Saliency}}, which is computed as:
\begin{equation}
\label{eq:ccs}
S_{\ell,t,i,j} = \max \big(0,\; \mu_c - \mu_b - \sigma_b \big),
\end{equation}
where $\mu_c$ and $\mu_b$ are the mean responses of $U_{\ell,t,i,j}$ computed over $K$ generated samples (we set $K=50$ to ensure statistical stability) under the concept and base prompts, respectively, and $\sigma_b$ is the standard deviation under the base prompts. 
Here, $\mu_c - \mu_b$ captures how strongly a neuron reacts to the target concept relative to the background, while subtracting $\sigma_b$ filters out neurons with unstable or noisy background activations. 
This ensures that only neurons with a stable and statistically significant increase for the target concept are retained, making $S_{\ell,t,i,j}$ a reliable indicator of concept-sensitive neurons.

\subsection{Concept-Sensitive Neurons}
\label{m:iden}

In this section, we introduce how to identify \emph{\textbf{Concept-Sensitive Neurons}}. These neurons are those most responsive to a specific target concept across both temporal and spatial contexts, ensuring that subsequent pruning focuses only on parameters that are truly concept-related.

\textbf{Time-Integrated Sensitivity.} 
Given the per-timestep contrastive concept saliency scores \(S_{\ell,t,i,j}\), we integrate temporal information to ensure robust identification of neurons across denoising steps.  
A direct summation across steps would overemphasize neurons that only spike briefly, so we design a two-part sensitivity measure balancing both \emph{response strength} and \emph{activation persistence}.  
For each neuron \((i,j)\), the aggregated sensitivity is
\begin{equation}
A_{\ell,i,j} = \tfrac{1}{2}\,\underbrace{\frac{1}{T}\sum_{t=1}^T S_{\ell,t,i,j}}_{\text{average response strength}} 
+ \tfrac{1}{2}\,\underbrace{\frac{1}{T}\sum_{t=1}^T \mathbf{1}[S_{\ell,t,i,j} > \tau_{\ell,t}]}_{\text{activation frequency}},
\end{equation}
where \(\mathbf{1}[\cdot]\) is the indicator function and \(T\) is the total number of denoising steps.  
The first term measures how strongly the neuron responds to the concept over the whole generation process, while the second term counts the proportion of timesteps where the neuron is significantly active.  
The threshold \(\tau_{\ell,t}\) is adaptively determined for each layer and timestep using a \emph{temporal sparsity} \(r_1\), by keeping the top-\(r_1\) fraction of neuron sensitivities:
\[
\tau_{\ell,t} = \mathrm{Top}\text{-}r_1\big(\{S_{\ell,t,i,j}\}_{i=1..C_{\mathrm{out}}, j=1..C_{\mathrm{in}}}\big),
\]
which dynamically adjusts to different noise levels and prevents a few extreme values from dominating the selection.  
By combining both terms with equal weight, neurons that are consistently and strongly relevant to the target concept are given higher \(A_{\ell,i,j}\).

\textbf{Spatial Sparsity Selection.} 
After obtaining the time-integrated sensitivity \(A_{\ell,i,j}\), we select neurons by jointly considering both channel-level and global-level information, ensuring that the most concept-relevant neurons are accurately identified.  

First, for each output channel \(i\), we rank all input neurons based on their sensitivities \(A_{\ell,i,j}\) and select the top \(k = r_2 \times C_{\mathrm{in}}\) neurons with the strongest responses, where \(r_2\) denotes the \emph{spatial sparsity}.  
This step focuses on identifying the most concept-relevant neurons for each channel individually, ensuring that the key neurons contributing to concept representation are preserved.  
The union of these channel-specific selections forms the local candidate set:
\begin{equation}
C_{\ell} = \bigcup_{i=1}^{C_{\mathrm{out}}} \bigl\{(i,j) \mid j \in \mathrm{Top}_k\{A_{\ell,i,j}\}_{j=1}^{C_{\mathrm{in}}}\bigr\}.
\end{equation}

Next, at the layer level, we consider all neurons across the entire layer and again rank them by their sensitivities.  
From this ranking, we retain the top \(K_{\mathrm{g}} = r_2 \times C_{\mathrm{out}} \times C_{\mathrm{in}}\) neurons with the highest response values.  
This global step filters out less relevant neurons and ensures that only the most strongly activated neurons across the whole layer are preserved, forming the global candidate set:
\begin{equation}
G_{\ell} = \bigl\{(i,j) \mid (i,j) \in \mathrm{Top}_{K_{\mathrm{g}}}\{A_{\ell,i,j}\}\bigr\}.
\end{equation}

Finally, we integrate the two selection criteria by taking their intersection:
\begin{equation}
\mathcal{Q}_{\ell}^{(c)} = C_{\ell} \cap G_{\ell}.
\end{equation}
This yields the final set of \emph{Concept-Sensitive Neurons} for layer \(\ell\) with respect to concept \(c\), guaranteeing that the selected neurons are globally competitive while remaining evenly distributed across channels.  
In other words, \(\mathcal{Q}_{\ell}^{(c)}\) captures the neurons that are most strongly associated with the target concept, both within their local context and across the entire network layer.

\subsection{Multi-concept Mask Fusion}
\label{m:fuse}

Building on the identified Concept-Sensitive Neurons \(\mathcal{Q}_{\ell}^{(c)}\), 
we generate a mask for each target concept:
\begin{equation}
\mathrm{Mask}_{\ell}^{(c)}(i,j)=
\begin{cases}
1, & \text{if } (i,j)\in \mathcal{Q}_{\ell}^{(c)},\\[4pt]
0, & \text{otherwise}.
\end{cases}
\end{equation}
By combining temporal aggregation and spatial sparsity into one unified decision, 
the resulting mask highlights only neurons that are consistently active and structurally significant 
for representing the target concept, while unrelated neurons are marked as 0 for potential pruning. 
This provides a stable and precise foundation for subsequent multi-concept unlearning.

A naïve mask-fusion strategy directly fuses the masks of all concepts and prunes any neuron that is sensitive to at least one of them.
In practice, this union approach severely degrades generation quality, since many neurons participate not only in specific concepts but also in core image formation and features. Empirically, we observe that a small subset of neurons responds strongly to \emph{most} or even \emph{all} of the concepts we wish to forget. Such neurons clearly encode broadly useful features rather
than any single concept, so we term these neurons \textbf{\emph{concept-agnostic neurons}}. 
To identify them, we compute each neuron’s aggregate concept sensitivity~(let \(C\) denote the total number of target concepts):
\begin{equation}
s_{\ell,i,j} = \sum_{c=1}^{C} \mathrm{Mask}^{(c)}_{\ell}[i,j],
\end{equation}
which counts how many of the concepts trigger a response in neuron \((i,j)\). 
We then define a \emph{concept-agnostic threshold}: 
\(\tau_{ca} = \lceil \alpha\,C\rceil\),
where the \textbf{\emph{concept-agnostic ratio}} \(\alpha\in(0,1]\) denotes the minimum fraction of concepts to which a neuron must be sensitive in order to be considered concept-agnostic.
During mask fusion, a neuron is labeled concept-agnostic and kept if $s_{\ell,i,j}\ge\tau_{ca}$.  
Consequently, only truly concept-sensitive neurons with $0< s_{\ell,i,j}<\tau_{ca}$ are pruned.
By preserving these concept-agnostic neurons, we retain the model's core generative capabilities and avoid the quality degradation caused by over-pruning.
We present comprehensive ablation results in Appendix~\ref{app:abl} to validate the effectiveness of preserving concept-agnostic neurons.

\section{Experiment}
\label{exp}
We first describe the experimental setup (Section~\ref{exp:setting}), followed by results on three unlearning tasks: object (Section~\ref{exp:object}), explicit content (Section~\ref{exp:nsfw}), and artistic style unlearning (Section~\ref{exp:art}), with comparisons to multiple baselines.
Additional experimental results and ablation studies are provided in Appendix~\ref{app:abl} and~\ref{app:res}.

\subsection{Experiment Setting}
\label{exp:setting}
All experiments were conducted using Stable Diffusion v1.5. To align with prior baselines, explicit content unlearning was evaluated on Stable Diffusion v1.4. To demonstrate the \emph{architectural generalizability} of our framework, we extend FIA to SDXL~\cite{podell2023sdxl} in Appendix~\ref{sec:sdxl_results}.
Multi-object unlearning is evaluated on the Imagenette benchmark~\citep{imagenette}, a ten-class ImageNet subset, while explicit content unlearning uses the I2P dataset~\citep{schramowski2023safe}, which includes diverse inappropriate prompts. For artistic style unlearning, we collected 200 artwork titles per artist and constructed prompts by appending the artist’s name. Model generative performance after unlearning is assessed on MS COCO-30K~\citep{lin2014microsoft}.
All experiments use 50 denoising steps. For Contrastive Concept Saliency, we generate 10 samples per concept, with a sensitivity analysis provided in Appendix~\ref{app:sample_size}.
Prompt design and FIA hyperparameter settings are given in Appendix~\ref{app:config}. All experiments were run on an NVIDIA RTX A6000 GPU, with peak GPU memory usage and execution time reported in Appendix~\ref{app:res} Table~\ref{tab:t} \ref{tab:t2}.

\captionsetup{font=footnotesize,labelfont=bf}
\begin{table*}[t]
  \centering
  \caption{Forgetting accuracy~($\downarrow$) for each class under simultaneous unlearning of ten concepts on Imagenette, and CLIP score ($\uparrow$). FIA (ours) achieves the best unlearning performance.}
  \label{tab:objects_1}
  
  \scriptsize 
  \renewcommand{\arraystretch}{1.1} 
  \setlength{\tabcolsep}{1pt}       

  \begin{tabularx}{\textwidth}{l *{9}{Y} >{\columncolor{gray!20}}Y}
    \toprule
    \multirow{2}{*}{\textbf{Imagenette classes}}
      & \multicolumn{10}{c}{\textbf{Method}} \\
    \cmidrule(lr){2-11}
    & SD v1.5
    & FMN
    & AC
    & ESD
    & SalUn
    & CP
    & MACE
    & UCE
    & SPM
    & \textbf{\emph{FIA}}~(Ours) \\
    \midrule
    \textbf{garbage truck}
      & 89.1 & 78.9 & 49.2 & 31.6 & 10.6 & 6.7  & 82.8 & 28.9 & 53.6 & \textbf{0.5} \\
    \textbf{cassette player}
      & 67.6 & 14.8 & 7.4  & 4.7  & 34.3 & 4.6  & 14.9 & 2.3  & 3.9  & \textbf{0.0} \\
    \textbf{tench}
      & 98.5 & 79.3 & 11.4 & 64.5 & 92.2 & 0.3  & 84.7 & 5.2  & 43.4 & \textbf{0.0} \\
    \textbf{English springer}
      & 98.2 & 85.6 & 92.1 & 79.3 & 1.5  & 2.5  & 93.1 & \textbf{0.8}  & 67.2 & 1.7  \\
    \textbf{chain saw}
      & 78.3 & 43.2 & 77.2 & 12.2 & 7.8  & \textbf{0.9}  & 73.5 & 4.7  & 32.7 & 1.8  \\
    \textbf{parachute}
      & 93.5 & 90.4 & 46.6 & 6.3  & 10.1 & 3.5  & 89.7 & 8.2  & 74.1 & \textbf{1.9} \\
    \textbf{golf ball}
      & 98.2 & 92.7 & 57.2 & 13.1 & 5.9  & 33.2 & 94.6 & 7.8  & 93.8 & \textbf{4.8} \\
    \textbf{church}
      & 86.9 & 77.5 & 82.9 & 61.4 & 1.2  & 8.1  & 69.3 & 19.5 & 66.5 & \textbf{0.0} \\
    \textbf{French horn}
      & 98.4 & 87.4 & 94.0 & 57.8 & 9.4  & \textbf{2.2}  & 96.4 & 3.6  & 17.3 & 2.9  \\
    \textbf{gas pump}
      & 94.7 & 69.1 & 63.5 & 54.9 & 58.7 & 11.4 & 83.2 & \textbf{5.2}  & 20.4 & 5.0  \\
    \midrule
    \textbf{Avg Acc} ($\downarrow$)
      & 90.34  & 71.89  & 58.15  & 38.58  & 23.17  &  7.34  & 78.22  &  8.62  & 47.29  & \textcolor{red}{\textbf{1.9}}  \\
    \textbf{CLIP$_{coco}$} ($\uparrow$)
      & 31.42  & 30.56  & \textbf{31.58}  & 30.12  & 29.93  & 27.93  & 31.05  & 29.25  & 30.77  & 29.56 \\
    \bottomrule
  \end{tabularx}
  \vspace{-1em}
\end{table*}

\captionsetup{font=footnotesize,labelfont=bf}
\begin{table*}[t]
  \centering
  \caption{Comparison of unlearning methods on Imagenette for simultaneous unlearning of the first five concepts. Reporting forgetting accuracy ($\downarrow$) on those five classes, preservation accuracy ($\uparrow$) on the last five, and harmonic-mean based overall score. Note that we do not highlight the best preserving accuracy in bold, as high values may result from failing to forget any concepts.}
  \label{tab:objects_2}
  
  \scriptsize 
  \renewcommand{\arraystretch}{1.1} 
  \setlength{\tabcolsep}{1pt}       

  \begin{tabularx}{\textwidth}{l *{8}{Y} >{\columncolor{gray!20}}Y}
    \toprule
    \multirow{2}{*}{\textbf{Imagenette classes}}
      & \multicolumn{9}{c}{\textbf{Method}} \\
    \cmidrule(lr){2-10}
    & FMN
    & AC
    & ESD
    & SalUn
    & CP
    & MACE
    & UCE
    & SPM
    & \textbf{\emph{FIA}}~(Ours) \\
    \midrule

    \multicolumn{1}{l}{\bfseries Classes to Forget}
      & \multicolumn{9}{c}{\bfseries Forgetting Accuracy ($\downarrow$)} \\
    \cmidrule(lr){1-10}
    \textbf{garbage truck}
      & 68.4 & 47.2 & 26.8 &  7.2 &  5.3 & 76.7 & 16.9 & 48.0 & \textbf{2.6} \\
    \textbf{cassette player}
      & 10.1 &  8.4 &  3.4 & 16.9 &  2.8 &  8.3 &  3.1 &  2.6 & \textbf{1.4} \\
    \textbf{tench}
      & 62.3 &  9.1 & 39.8 & 43.1 &  1.9 & 65.5 &  3.7 & 46.2 & \textbf{1.7} \\
    \textbf{English springer}
      & 79.6 & 76.4 & 53.7 &  1.3 &  0.9 & 80.6 & \textbf{0.4} & 59.8 &  2.5 \\
    \textbf{chain saw}
      & 23.9 & 64.2 &  9.5 &  8.1 &  2.8 & 61.3 &  3.6 & 28.2 & \textbf{2.4} \\

    \midrule
    \multicolumn{1}{l}{\bfseries Classes to Preserve}
      & \multicolumn{9}{c}{\bfseries Preserving Accuracy ($\uparrow$)} \\
    \cmidrule(lr){1-10}
    \textbf{parachute}
      & 79.2 & 65.1 & 71.0 & 73.3 & 48.3 & 80.7 & 58.6 & 77.1 & 77.2 \\
    \textbf{golf ball}
      & 85.8 & 77.4 & 74.7 & 81.4 & 62.7 & 81.2 & 76.8 & 85.1 & 81.7 \\
    \textbf{church}
      & 71.4 & 79.0 & 63.9 & 72.4 & 51.0 & 66.8 & 75.0 & 68.5 & 68.9 \\
    \textbf{French horn}
      & 80.7 & 90.1 & 87.0 & 85.5 & 84.0 & 87.3 & 78.7 & 80.3 & 86.4 \\
    \textbf{gas pump}
      & 67.4 & 78.9 & 64.4 & 74.2 & 22.5 & 74.9 & 75.2 & 73.0 & 67.9 \\

    \midrule
    \textbf{Forgetting Acc [1--5]} ($\downarrow$)
      & 48.9 & 41.1 & 26.6 & 22.3 &  2.7 & 58.5 &  5.5 & 37.0 & \textbf{2.1} \\
    \textbf{Preserving Acc [6--10]} ($\uparrow$)
      & 76.9 & 78.1 & 72.2 & 77.4 & 52.4 & \textbf{78.2} & 71.9 & 76.5 & 76.7 \\
    \textbf{Overall Score} ($\uparrow$)
      & 61.4 & 67.2 & 72.8 & 77.5 & 68.1 & 54.2 & 81.7 & 69.1 & \textcolor{red}{\textbf{86.0}} \\

    \bottomrule
  \end{tabularx}
  \vspace{-1em}
\end{table*}

\begin{figure*}[t]
  \centering
  \includegraphics[width=1\textwidth]{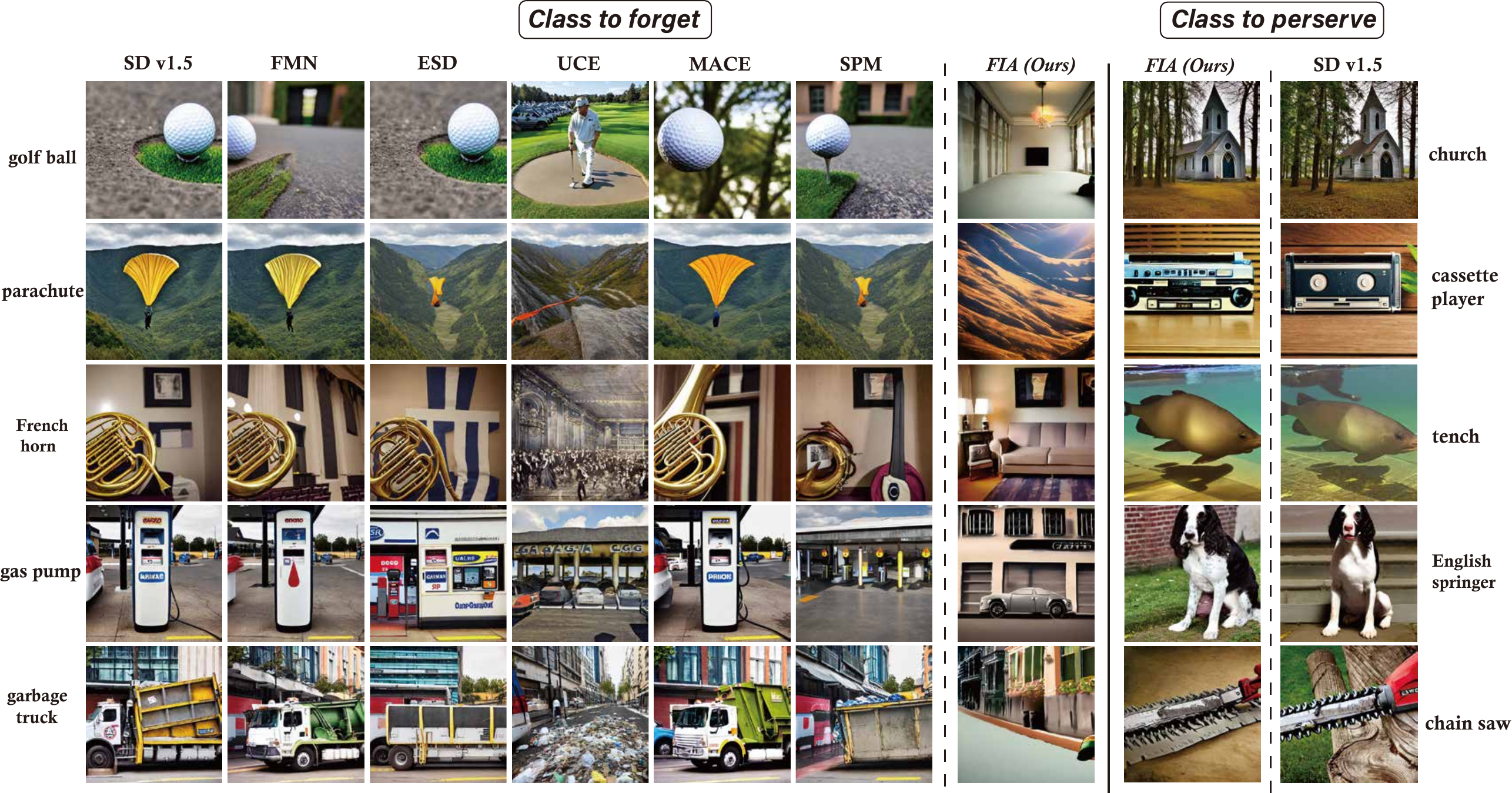}
  \caption{Visual results on the Imagenette dataset, demonstrating simultaneous unlearning of five target classes while preserving the other five. Our method achieves superior unlearning performance on the target classes, and continues to faithfully generate the preserved classes. More visual results can be found in Figure~\ref{app:obj1},\ref{app:obj2}.}
  \label{fig:obj1}
  \vspace{-0.5em}
\end{figure*}

\begin{figure*}[t]
  \centering
  \includegraphics[width=1\textwidth]{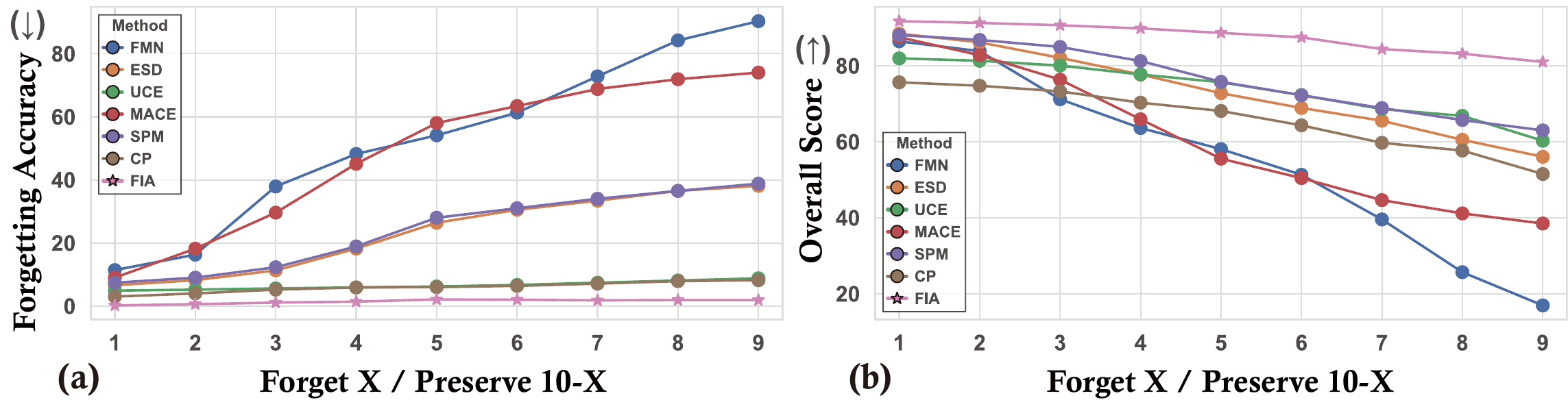}
  \caption{Forgetting accuracy (a) and Overall Score (b) on Imagenette across various forget–preserve configurations, demonstrating FIA’s superior balance between unlearning efficacy and generation quality.}
  \label{fig:obj2}
  \vspace{-1em}
\end{figure*}

\subsection{Multi-object Unlearning}  
\label{exp:object}

In Table~\ref{tab:objects_1}, we report results for simultaneously forgetting all ten Imagenette classes. Forgetting accuracy is measured using a pretrained ResNet-50 classifier~\citep{he2016deep}, following the same evaluation protocol as prior work. FIA achieves the lowest average forgetting accuracy of 1.9\%, indicating the most complete removal of target concepts.  
In terms of generative quality, FIA attains a CLIP score of 29.56 on MS COCO 30K~\citep{lin2014microsoft}, outperforming training-free baselines such as CP~\citep{chavhan2024conceptprune} and UCE~\cite{gandikota2024unified}. While several finetuning-based methods (FMN~\cite{zhang2024forget}, ESD~\citep{gandikota2023erasing}, AC~\citep{kumari2023ablating}, SalUn~\citep{fan2023salun}, SPM~\citep{lyu2024one}, and MACE~\citep{lu2024mace}) achieve slightly higher CLIP scores, they suffer from substantially higher forgetting accuracies. In contrast, FIA delivers markedly lower forgetting while maintaining competitive generative performance, demonstrating superior concept unlearning.
In Table~\ref{tab:objects_2}, we further evaluate the trade-off between forgetting five classes and preserving the remaining five. We intentionally avoid bolding the highest preserving accuracy, as models that fail to forget may trivially retain all concepts and obtain inflated scores. Instead, we report an \emph{Overall Score} that jointly reflects forgetting and preservation. This score is defined as the harmonic mean of preserving accuracy \(P\) and forgetting rate \(R = 1 - F\), i.e.\ \(\mathrm{Overall\ Score} = 2\,\frac{P(1 - F)}{P + (1 - F)} \times 100\%\). FIA achieves 86\% on this metric, outperforming all baselines. Overall, these results confirm that FIA achieves the best balance between effective concept unlearning and preservation quality, establishing a new state of the art for multi-concept unlearning. Additional results are provided in Appendix~\ref{app:res} (Figure~\ref{app:obj1}--\ref{app:obj2} and Table~\ref{tab:objects_st1}--\ref{tab:objects_st2}).

To systematically evaluate both unlearning efficacy and concept preservation, we erase $x$ Imagenette classes and preserve the remaining ones, measuring how well each method erases target concepts while maintaining other content. As shown in Figure~\ref{fig:obj2}(a), FIA consistently yields the lowest forgetting accuracy for every forget–preserve configuration, outperforming all baselines even as the number of forgotten concepts increases. The overall score plotted in Figure~\ref{fig:obj2}(b) further demonstrates that FIA achieves the best trade-off between concept removal and generation quality. These results confirm the robustness of our approach in balancing forgetting and preservation across all settings. We also \emph{expand multi-concept unlearning to a larger scale}, and we show the relevant results in Appendix~\ref{app:res} (Figure~\ref{app:s} and Table~\ref{tab:s}). 

\begin{table*}[t]
  \centering
  \caption{Results of NudeNet detection result on the I2P dataset. “(F)” denotes female, “(M)” denotes male. $^{\dagger}$ Partial results from MACE~\citep{lu2024mace} and SA~\citep{heng2023selective}.}
  {\scriptsize
    \setlength{\tabcolsep}{3pt}
    \resizebox{\textwidth}{!}{%
      \begin{tabular}{@{}lccccccccc|c|cc@{}}
        \toprule
        \multirow{2}{*}{\textbf{Method}}
          & \multicolumn{9}{c}{\textbf{NudeNet Detection}}
          & \multicolumn{2}{c}{\textbf{Metric}} \\
        \cmidrule(lr){2-10}\cmidrule(lr){11-12}
          & \textbf{Armpits} & \textbf{Belly} & \textbf{Buttocks} & \textbf{Feet}
          & \textbf{Breasts (F)} & \textbf{Genitalia (F)}
          & \textbf{Breasts (M)} & \textbf{Genitalia (M)}
          & \textbf{Total} $\downarrow$ & \textbf{FID} $\downarrow$ & \textbf{CLIP} $\uparrow$ \\
        \midrule
        FMN
          &  47  & 120  &  23  &  54  & 163  &  17  &  21  &   3   & 448   & 13.54 & 30.43 \\
        AC
          & 153  & 180  &  45  &  66  & 298  &  22  &  67  &   7   & 838   & 14.13 & \textbf{31.37} \\
        UCE
          &  29  &  62  &   7  &  29  &  35  &   5  &  11  &   4   & 182   & 14.07 & 30.85 \\
        CP
          &  36  &  31  &   7  &   8  &  49  &   4  &   4  &   9   & 148   & 14.11 & 31.04 \\
        SLD-M
          &  47  &  72  &   3  &  21  &  39  &   1  &  26  &   3   & 212   & 16.34 & 30.90 \\
        ESD
          &  59  &  73  &  12  &  39  & 100  &   6  &  18  &   8   & 315   & 14.41 & 30.69 \\
        SA$^{\dagger}$
          &  72  &  77  &  19  &  25  &  83  &  16  &   \textbf{0}  &   \textbf{0}   & 292   & --     & --     \\
        SPM
          &  51  &  69  &   8  &  14  &  70  &   5  &  10  &   2   & 229   & 13.81 & 31.24 \\
        MACE$^{\dagger}$
          &  17  &  19  & \textbf{2}  &  39  &  16  &   2  &   9  &   7   & 111 & \textbf{13.42} & 29.41 \\
        \rowcolor{gray!20}
        \textbf{\emph{FIA}} (Ours)
          & \textbf{6}   & \textbf{2}   &   7  & \textbf{2}   & \textbf{6}   & \textbf{0}   &  1   &   8   &  \textcolor{red}{\textbf{32}}  & 14.02 & 31.18 \\
        \rowcolor{white}
        \midrule
        SD v1.4
          & 148  & 170  &  29  &  63  & 266  &  18  &  42  &   7   & 743   & 14.04 & 31.34 \\
        \bottomrule
      \end{tabular}%
    }
  }
  \label{tab:nsfw}
\end{table*}

\begin{figure*}[t]
  \centering
  \begin{minipage}[t]{0.48\textwidth}
    \captionof{table}{Comparison of unlearning methods for simultaneous unlearning of five artist styles.}
    \label{tab:art_1}
    \resizebox{\textwidth}{!}{%
      \renewcommand{\arraystretch}{0.9}
      \small
      \begin{tabular}{lccccc}
        \toprule
        \multirow{2}{*}{\textbf{Method}}
          & \multicolumn{2}{c}{\textbf{Artist unlearning}}
          & \multicolumn{2}{c}{\textbf{MS COCO-30K}}
          & \multirow{2}{*}{\textbf{Rank $\downarrow$}} \\
        \cmidrule(lr){2-3}\cmidrule(lr){4-5}
          & \textbf{CLIP$_a$ ↓} & \textbf{FSR ↑} & \textbf{FID ↓} & \textbf{CLIP ↑} & {} \\
        \midrule
        FMN & 30.27 & 52.8 & 21.4 & 30.82 & 4.75 \\
        ESD & 33.62 & 39.2 & 17.1 & 30.52 & 6.50 \\
        UCE & 31.89 & 44.0 & 19.7 & 31.19 & 5.50 \\
        AC  & 33.59 & 45.2 & 16.6 & 31.28 & 4.00 \\
        CP  & 27.90 & 79.6 & 18.4 & 29.76 & 4.50 \\
        MACE& 30.98 & 57.4 & 15.9 & 30.14 & 3.75 \\
        SPM & 31.10 & 40.0 & 17.4 & 31.33 & 4.50 \\
        \rowcolor{gray!20}
        \emph{FIA} (Ours) & \textbf{27.45} & \textbf{83.4} & 16.7 & 30.56 & \textcolor{red}{\textbf{2.50}} \\
        \midrule
        SD v1.5 & 42.10 & -- & 14.5 & 31.34 & -- \\
        \bottomrule
      \end{tabular}
    }
  \end{minipage}\hfill
  \begin{minipage}[t]{0.48\textwidth}
    \captionof{table}{Comparison of per-style CLIP scores for artist style unlearning.}
    \label{tab:art_2}
    \resizebox{\textwidth}{!}{%
      \renewcommand{\arraystretch}{1.15}
      \large
      \begin{tabular}{@{}lccccc|c@{}}
        \toprule
        \multirow{2}{*}{\textbf{Method}}
          & \multicolumn{5}{c}{\textbf{Artist Style}}
          & \multirow{2}{*}{\textbf{Avg CLIP} $\downarrow$} \\
        \cmidrule(lr){2-6}
          & \textbf{Van Gogh} & \textbf{Monet} & \textbf{Picasso} & \textbf{Da Vinci} & \textbf{Dali} & {} \\
        \midrule
        FMN & 32.67 & 32.92 & 28.04 & 28.89 & 28.83 & 30.27 \\
        ESD & 34.41 & 35.39 & 31.89 & 33.17 & 33.24 & 33.62 \\
        UCE & 33.98 & 34.49 & 30.16 & 29.99 & 30.83 & 31.89 \\
        AC  & 33.40 & 34.30 & 33.30 & 33.80 & 33.15 & 33.59 \\
        CP  & 27.94 & 25.33 & 28.01 & 28.15 & 29.67 & 27.90 \\
        MACE& 31.56 & 33.91 & 31.34 & 29.01 & 29.08 & 30.98 \\
        SPM & 32.55 & 33.66 & 29.20 & 29.29 & 30.80 & 31.10 \\
        \rowcolor{gray!20}
        \emph{FIA} (Ours) & 28.13 & 26.37 & 27.16 & 27.44 & 28.15 & \textcolor{red}{\textbf{27.45}} \\
        \bottomrule
      \end{tabular}
    }
  \end{minipage}
\end{figure*}

\subsection{Explicit Content Unlearning}
\label{exp:nsfw}


We evaluate explicit concept unlearning on the I2P (“Inappropriate Image Prompts”) benchmark, which contains 4,703 real-world text-to-image prompts that are highly prone to generate inappropriate content.
Our evaluation uses three complementary metrics: (1) NudeNet detection counts to quantify concept unlearning, (2) FID~(Fréchet Inception Distance) to assess visual quality, and (3) CLIP score to measure semantic consistency.  NudeNet~\citep{bedapudi2019nudenet} is a lightweight nudity detector offering both classification and localization of explicit elements to identify exposed body regions (armpits, belly, buttocks, feet, female breasts, female genitalia, male breasts, and male genitalia). 
Following prior work, we mark a region as inappropriate only if NudeNet’s confidence exceeds 0.6, matching all baselines’ settings~\citep{schramowski2023safe, lu2024mace}. 
As shown in Table~\ref{tab:nsfw}, FIA reduces the total number of NudeNet detections on Stable Diffusion v1.4 from 743 to just 32, outperforming every competitor. 
On MS COCO 30K~\cite{lin2014microsoft}, FIA achieves an FID of 14.02, nearly matching MACE’s best score of 13.42, and a CLIP score of 31.18, demonstrating that images after unlearning maintain high visual quality and semantic fidelity. These results establish FIA as the new state of the art for explicit concept unlearning under strict, reproducible conditions.
More experimental and visual results are presented in Appendix~\ref{app:res} (Figure~\ref{app:nsfw} and Table~\ref{tab:nsfw_2}).

\subsection{Multi-Artistic-Style Unlearning}
\label{exp:art}


For artistic styles unlearning, our goal is to simultaneously forget the styles of five famous artists (Van Gogh, Monet, Picasso, Da Vinci, and Dali).  
To evaluate unlearning effectiveness, we report the CLIP score (\(\mathrm{CLIP}_a\)) and \textbf{\emph{Forget-Success Rate}}~(FSR). 
For each prompt we generate paired images with the original SD model and with the edited model, using the same random seed. we count a success whenever the edited image yields a lower CLIP score than the original.  
Then FSR is defined as:
{\small
\begin{equation}
\mathrm{FSR} = \frac{1}{N}\sum_{i=1}^{N}\mathbf{1}\bigl(\mathrm{CLIP}_a^{(i),\mathrm{edited}}<\mathrm{CLIP}_a^{(i),\mathrm{orig}}\bigr)
\end{equation}
}
where $N$ is the total number of prompts. 
We assess image quality after unlearning on MS COCO-30K~\citep{lin2014microsoft} via FID and CLIP score.
To combine these four metrics into one measure, we compute each method’s \textbf{\emph{average rank}} over these four measures, with a lower rank indicating better overall performance. 
As shown in Table~\ref{tab:art_1} and Table~\ref{tab:art_2}, FIA achieves the best average rank, showcasing its superior unlearning capability and the optimal trade-off between unlearning effectiveness and image quality. More results can be found in Appendix~\ref{app:res} (Figure~\ref{app:art} and Table~\ref{tab:art_st}).

\vspace{-0.8em}
\section{Conclusion}
\label{conclusion}
FIA offers a simple, training-free solution for removing multiple unwanted concepts from diffusion models. By pruning fewer than 0.3\% of neurons, FIA achieves state-of-the-art unlearning performance while preserving high-quality, semantically faithful image generation without additional fine-tuning. As the number of target concepts scales into the hundreds, the required sparsity inevitably increases, leading to degradation in image quality. Future work will integrate FIA with fine-tuning techniques to address this limitation and better preserve generation quality at larger unlearning scales. In conclusion, this work resolves the core trade-off between effective concept unlearning and generation quality, offering a lightweight approach suited for real-world privacy, copyright, and safety requirements. FIA lays the foundation for more controllable and secure generative models.

\clearpage
\newpage
\section*{Impact Statement}
This work introduces a training-free unlearning framework designed to remove specific concepts from text-to-image diffusion models. 
The primary societal impact of our method is positive: it provides a practical tool for adhering to \textbf{copyright regulations}, allowing model owners to remove protected artistic styles or characters without prohibitive retraining costs. 
Furthermore, it enhances \textbf{AI safety} by effectively erasing harmful, explicit, or biased content from open-source models, thereby mitigating the risk of generating non-consensual or inappropriate imagery.

\bibliography{cited_paper}

@article{kazerouni2022diffusion,
  title={Diffusion models for medical image analysis: A comprehensive survey},
  author={Kazerouni, Amirhossein and Aghdam, Ehsan Khodapanah and Heidari, Moein and Azad, Reza and Fayyaz, Mohsen and Hacihaliloglu, Ilker and Merhof, Dorit},
  journal={arXiv preprint arXiv:2211.07804},
  year={2022}
}

@article{xing2024survey,
  title={A survey on video diffusion models},
  author={Xing, Zhen and Feng, Qijun and Chen, Haoran and Dai, Qi and Hu, Han and Xu, Hang and Wu, Zuxuan and Jiang, Yu-Gang},
  journal={ACM Computing Surveys},
  volume={57},
  number={2},
  pages={1--42},
  year={2024},
  publisher={ACM New York, NY}
}

@article{jang2022knowledge,
  title={Knowledge unlearning for mitigating privacy risks in language models},
  author={Jang, Joel and Yoon, Dongkeun and Yang, Sohee and Cha, Sungmin and Lee, Moontae and Logeswaran, Lajanugen and Seo, Minjoon},
  journal={arXiv preprint arXiv:2210.01504},
  year={2022}
}

@article{nguyen2022survey,
  title={A survey of machine unlearning},
  author={Nguyen, Thanh Tam and Huynh, Thanh Trung and Ren, Zhao and Nguyen, Phi Le and Liew, Alan Wee-Chung and Yin, Hongzhi and Nguyen, Quoc Viet Hung},
  journal={arXiv preprint arXiv:2209.02299},
  year={2022}
}

@article{weidinger2021ethical,
  title={Ethical and social risks of harm from language models},
  author={Weidinger, Laura and Mellor, John and Rauh, Maribeth and Griffin, Conor and Uesato, Jonathan and Huang, Po-Sen and Cheng, Myra and Glaese, Mia and Balle, Borja and Kasirzadeh, Atoosa and others},
  journal={arXiv preprint arXiv:2112.04359},
  year={2021}
}

@article{brundage2018malicious,
  title={The malicious use of artificial intelligence: Forecasting, prevention, and mitigation},
  author={Brundage, Miles and Avin, Shahar and Clark, Jack and Toner, Helen and Eckersley, Peter and Garfinkel, Ben and Dafoe, Allan and Scharre, Paul and Zeitzoff, Thomas and Filar, Bobby and others},
  journal={arXiv preprint arXiv:1802.07228},
  year={2018}
}

@article{tolosana2020deepfakes,
  title={Deepfakes and beyond: A survey of face manipulation and fake detection},
  author={Tolosana, Ruben and Vera-Rodriguez, Ruben and Fierrez, Julian and Morales, Aythami and Ortega-Garcia, Javier},
  journal={Information Fusion},
  volume={64},
  pages={131--148},
  year={2020},
  publisher={Elsevier}
}

@article{solaiman2021process,
  title={Process for adapting language models to society (palms) with values-targeted datasets},
  author={Solaiman, Irene and Dennison, Christy},
  journal={Advances in Neural Information Processing Systems},
  volume={34},
  pages={5861--5873},
  year={2021}
}

@inproceedings{rombach2022high,
  title={High-resolution image synthesis with latent diffusion models},
  author={Rombach, Robin and Blattmann, Andreas and Lorenz, Dominik and Esser, Patrick and Ommer, Bj{\"o}rn},
  booktitle={Proceedings of the IEEE/CVF conference on computer vision and pattern recognition},
  pages={10684--10695},
  year={2022}
}

@inproceedings{ma2024sit,
  title={Sit: Exploring flow and diffusion-based generative models with scalable interpolant transformers},
  author={Ma, Nanye and Goldstein, Mark and Albergo, Michael S and Boffi, Nicholas M and Vanden-Eijnden, Eric and Xie, Saining},
  booktitle={European Conference on Computer Vision},
  pages={23--40},
  year={2024},
  organization={Springer}
}

@article{podell2023sdxl,
  title={Sdxl: Improving latent diffusion models for high-resolution image synthesis},
  author={Podell, Dustin and English, Zion and Lacey, Kyle and Blattmann, Andreas and Dockhorn, Tim and M{\"u}ller, Jonas and Penna, Joe and Rombach, Robin},
  journal={arXiv preprint arXiv:2307.01952},
  year={2023}
}

@article{saharia2022photorealistic,
  title={Photorealistic text-to-image diffusion models with deep language understanding},
  author={Saharia, Chitwan and Chan, William and Saxena, Saurabh and Li, Lala and Whang, Jay and Denton, Emily L and Ghasemipour, Kamyar and Gontijo Lopes, Raphael and Karagol Ayan, Burcu and Salimans, Tim and others},
  journal={Advances in neural information processing systems},
  volume={35},
  pages={36479--36494},
  year={2022}
}

@article{nichol2021glide,
  title={Glide: Towards photorealistic image generation and editing with text-guided diffusion models},
  author={Nichol, Alex and Dhariwal, Prafulla and Ramesh, Aditya and Shyam, Pranav and Mishkin, Pamela and McGrew, Bob and Sutskever, Ilya and Chen, Mark},
  journal={arXiv preprint arXiv:2112.10741},
  year={2021}
}

@article{ding2022cogview2,
  title={Cogview2: Faster and better text-to-image generation via hierarchical transformers},
  author={Ding, Ming and Zheng, Wendi and Hong, Wenyi and Tang, Jie},
  journal={Advances in Neural Information Processing Systems},
  volume={35},
  pages={16890--16902},
  year={2022}
}

@inproceedings{zhou2022towards,
  title={Towards language-free training for text-to-image generation},
  author={Zhou, Yufan and Zhang, Ruiyi and Chen, Changyou and Li, Chunyuan and Tensmeyer, Chris and Yu, Tong and Gu, Jiuxiang and Xu, Jinhui and Sun, Tong},
  booktitle={Proceedings of the IEEE/CVF conference on computer vision and pattern recognition},
  pages={17907--17917},
  year={2022}
}

@inproceedings{bourtoule2021machine,
  title={Machine unlearning},
  author={Bourtoule, Lucas and Chandrasekaran, Varun and Choquette-Choo, Christopher A and Jia, Hengrui and Travers, Adelin and Zhang, Baiwu and Lie, David and Papernot, Nicolas},
  booktitle={2021 IEEE symposium on security and privacy (SP)},
  pages={141--159},
  year={2021},
  organization={IEEE}
}

@article{guo2019certified,
  title={Certified data removal from machine learning models},
  author={Guo, Chuan and Goldstein, Tom and Hannun, Awni and Van Der Maaten, Laurens},
  journal={arXiv preprint arXiv:1911.03030},
  year={2019}
}

@inproceedings{graves2021amnesiac,
  title={Amnesiac machine learning},
  author={Graves, Laura and Nagisetty, Vineel and Ganesh, Vijay},
  booktitle={Proceedings of the AAAI Conference on Artificial Intelligence},
  volume={35},
  number={13},
  pages={11516--11524},
  year={2021}
}

@article{vatter2023evolution,
  title={The evolution of distributed systems for graph neural networks and their origin in graph processing and deep learning: A survey},
  author={Vatter, Jana and Mayer, Ruben and Jacobsen, Hans-Arno},
  journal={ACM Computing Surveys},
  volume={56},
  number={1},
  pages={1--37},
  year={2023},
  publisher={ACM New York, NY, USA}
}

@article{xu2024machine,
  title={Machine unlearning: Solutions and challenges},
  author={Xu, Jie and Wu, Zihan and Wang, Cong and Jia, Xiaohua},
  journal={IEEE Transactions on Emerging Topics in Computational Intelligence},
  year={2024},
  publisher={IEEE}
}

@article{marino2025bridge,
  title={Bridge the Gaps between Machine Unlearning and AI Regulation},
  author={Marino, Bill and Kurmanji, Meghdad and Lane, Nicholas D},
  journal={arXiv preprint arXiv:2502.12430},
  year={2025}
}

@article{zhang2023review,
  title={A review on machine unlearning},
  author={Zhang, Haibo and Nakamura, Toru and Isohara, Takamasa and Sakurai, Kouichi},
  journal={SN Computer Science},
  volume={4},
  number={4},
  pages={337},
  year={2023},
  publisher={Springer}
}

@article{liu2024machine,
  title={Machine unlearning in generative ai: A survey},
  author={Liu, Zheyuan and Dou, Guangyao and Tan, Zhaoxuan and Tian, Yijun and Jiang, Meng},
  journal={arXiv preprint arXiv:2407.20516},
  year={2024}
}

@article{jia2023model,
  title={Model sparsity can simplify machine unlearning},
  author={Jia, Jinghan and Liu, Jiancheng and Ram, Parikshit and Yao, Yuguang and Liu, Gaowen and Liu, Yang and Sharma, Pranay and Liu, Sijia},
  journal={Advances in Neural Information Processing Systems},
  volume={36},
  pages={51584--51605},
  year={2023}
}

@inproceedings{kumari2023ablating,
  title={Ablating concepts in text-to-image diffusion models},
  author={Kumari, Nupur and Zhang, Bingliang and Wang, Sheng-Yu and Shechtman, Eli and Zhang, Richard and Zhu, Jun-Yan},
  booktitle={Proceedings of the IEEE/CVF International Conference on Computer Vision},
  pages={22691--22702},
  year={2023}
}

@article{chavhan2024conceptprune,
  title={ConceptPrune: Concept editing in diffusion models via skilled neuron pruning},
  author={Chavhan, Ruchika and Li, Da and Hospedales, Timothy},
  journal={arXiv preprint arXiv:2405.19237},
  year={2024}
}

@inproceedings{gandikota2023erasing,
  title={Erasing concepts from diffusion models},
  author={Gandikota, Rohit and Materzynska, Joanna and Fiotto-Kaufman, Jaden and Bau, David},
  booktitle={Proceedings of the IEEE/CVF International Conference on Computer Vision},
  pages={2426--2436},
  year={2023}
}

@inproceedings{zhang2024forget,
  title={Forget-me-not: Learning to forget in text-to-image diffusion models},
  author={Zhang, Gong and Wang, Kai and Xu, Xingqian and Wang, Zhangyang and Shi, Humphrey},
  booktitle={Proceedings of the IEEE/CVF conference on computer vision and pattern recognition},
  pages={1755--1764},
  year={2024}
}

@inproceedings{schramowski2023safe,
  title={Safe latent diffusion: Mitigating inappropriate degeneration in diffusion models},
  author={Schramowski, Patrick and Brack, Manuel and Deiseroth, Bj{\"o}rn and Kersting, Kristian},
  booktitle={Proceedings of the IEEE/CVF Conference on Computer Vision and Pattern Recognition},
  pages={22522--22531},
  year={2023}
}

@article{fan2023salun,
  title={Salun: Empowering machine unlearning via gradient-based weight saliency in both image classification and generation},
  author={Fan, Chongyu and Liu, Jiancheng and Zhang, Yihua and Wong, Eric and Wei, Dennis and Liu, Sijia},
  journal={arXiv preprint arXiv:2310.12508},
  year={2023}
}

@article{heng2023selective,
  title={Selective amnesia: A continual learning approach to forgetting in deep generative models},
  author={Heng, Alvin and Soh, Harold},
  journal={Advances in Neural Information Processing Systems},
  volume={36},
  pages={17170--17194},
  year={2023}
}

@inproceedings{lu2024mace,
  title={Mace: Mass concept erasure in diffusion models},
  author={Lu, Shilin and Wang, Zilan and Li, Leyang and Liu, Yanzhu and Kong, Adams Wai-Kin},
  booktitle={Proceedings of the IEEE/CVF Conference on Computer Vision and Pattern Recognition},
  pages={6430--6440},
  year={2024}
}

@inproceedings{lyu2024one,
  title={One-dimensional adapter to rule them all: Concepts diffusion models and erasing applications},
  author={Lyu, Mengyao and Yang, Yuhong and Hong, Haiwen and Chen, Hui and Jin, Xuan and He, Yuan and Xue, Hui and Han, Jungong and Ding, Guiguang},
  booktitle={Proceedings of the IEEE/CVF Conference on Computer Vision and Pattern Recognition},
  pages={7559--7568},
  year={2024}
}

@article{zhao2024separable,
  title={Separable multi-concept erasure from diffusion models},
  author={Zhao, Mengnan and Zhang, Lihe and Zheng, Tianhang and Kong, Yuqiu and Yin, Baocai},
  journal={arXiv preprint arXiv:2402.05947},
  year={2024}
}

@inproceedings{gandikota2024unified,
  title={Unified concept editing in diffusion models},
  author={Gandikota, Rohit and Orgad, Hadas and Belinkov, Yonatan and Materzy{\'n}ska, Joanna and Bau, David},
  booktitle={Proceedings of the IEEE/CVF Winter Conference on Applications of Computer Vision},
  pages={5111--5120},
  year={2024}
}

@article{deng2026forget,
  title={Forget Many, Forget Right: Scalable and Precise Concept Unlearning in Diffusion Models},
  author={Deng, Kaiyuan and Li, Gen and Xiao, Yang and Hui, Bo and Ma, Xiaolong},
  journal={arXiv preprint arXiv:2601.06162},
  year={2026}
}

@inproceedings{li2025sculpting,
  title={Sculpting memory: Multi-concept forgetting in diffusion models via dynamic mask and concept-aware optimization},
  author={Li, Gen and Xiao, Yang and Ji, Jie and Deng, Kaiyuan and Hui, Bo and Guo, Linke and Ma, Xiaolong},
  booktitle={Proceedings of the IEEE/CVF International Conference on Computer Vision},
  pages={19659--19668},
  year={2025}
}

@article{xiao2025right,
  title={The Right to be Forgotten in Pruning: Unveil Machine Unlearning on Sparse Models},
  author={Xiao, Yang and Li, Gen and Ji, Jie and Ye, Ruimeng and Ma, Xiaolong and Hui, Bo},
  journal={arXiv preprint arXiv:2507.18725},
  year={2025}
}

@inproceedings{xiao2025efficient,
  title={Efficient Knowledge Graph Unlearning with Zeroth-order Information},
  author={Xiao, Yang and Ye, Ruimeng and Liu, Bohan and Ma, Xiaolong and Hui, Bo},
  booktitle={Proceedings of the 34th ACM International Conference on Information and Knowledge Management},
  pages={3509--3519},
  year={2025}
}

@inproceedings{li2024neurrev,
  title={Neurrev: Train better sparse neural network practically via neuron revitalization},
  author={Li, Gen and Yin, Lu and Ji, Jie and Niu, Wei and Qin, Minghai and Ren, Bin and Guo, Linke and Liu, Shiwei and Ma, Xiaolong},
  booktitle={The Twelfth International Conference on Learning Representations},
  year={2024}
}

@article{xiao2025lightcache,
  title={LightCache: Memory-Efficient, Training-Free Acceleration for Video Generation},
  author={Xiao, Yang and Li, Gen and Deng, Kaiyuan and Wu, Yushu and Zhan, Zheng and Wang, Yanzhi and Ma, Xiaolong and Hui, Bo},
  journal={arXiv preprint arXiv:2510.05367},
  year={2025}
}

@inproceedings{srivatsan2025stereo,
  title={Stereo: A two-stage framework for adversarially robust concept erasing from text-to-image diffusion models},
  author={Srivatsan, Koushik and Shamshad, Fahad and Naseer, Muzammal and Patel, Vishal M and Nandakumar, Karthik},
  booktitle={Proceedings of the IEEE/CVF Conference on Computer Vision and Pattern Recognition},
  pages={23765--23774},
  year={2025}
}

@article{li2025speed,
  title={Speed: Scalable, precise, and efficient concept erasure for diffusion models},
  author={Li, Ouxiang and Wang, Yuan and Hu, Xinting and Jiang, Houcheng and Liang, Tao and Hao, Yanbin and Ma, Guojun and Feng, Fuli},
  journal={arXiv preprint arXiv:2503.07392},
  year={2025}
}

@misc{imagenette,
  author = {Howard, Jeremy and others},
  title = {Imagenette: A smaller subset of ImageNet for quick experiments},
  howpublished = {\url{https://github.com/fastai/imagenette}},
  year = {2019}
}

@inproceedings{lin2014microsoft,
  title={Microsoft coco: Common objects in context},
  author={Lin, Tsung-Yi and Maire, Michael and Belongie, Serge and Hays, James and Perona, Pietro and Ramanan, Deva and Doll{\'a}r, Piotr and Zitnick, C Lawrence},
  booktitle={Computer vision--ECCV 2014: 13th European conference, zurich, Switzerland, September 6-12, 2014, proceedings, part v 13},
  pages={740--755},
  year={2014},
  organization={Springer}
}

@inproceedings{he2016deep,
  title={Deep residual learning for image recognition},
  author={He, Kaiming and Zhang, Xiangyu and Ren, Shaoqing and Sun, Jian},
  booktitle={Proceedings of the IEEE conference on computer vision and pattern recognition},
  pages={770--778},
  year={2016}
}

@misc{bedapudi2019nudenet,
  author       = {Bedapudi, Praneeth},
  title        = {NudeNet: An ensemble of Neural Nets for Nudity Detection and Censoring},
  howpublished = {Medium blog post},
  year         = {2019},
  month        = mar
}

@article{lee2018snip,
  title={Snip: Single-shot network pruning based on connection sensitivity},
  author={Lee, Namhoon and Ajanthan, Thalaiyasingam and Torr, Philip HS},
  journal={arXiv preprint arXiv:1810.02340},
  year={2018}
}

@article{wang2020picking,
  title={Picking winning tickets before training by preserving gradient flow},
  author={Wang, Chaoqi and Zhang, Guodong and Grosse, Roger},
  journal={arXiv preprint arXiv:2002.07376},
  year={2020}
}
\bibliographystyle{icml2026}

\newpage
\appendix
\onecolumn
\section{Detailed Configurations for FIA}
\label{app:config}

\subsection{Prompt Design}
\label{app:prompt}
In this section, we first provide detailed information on the prompt designs and the hyperparameter settings of our framework. Table~\ref{tab:pconig1} shows the prompt templates used in our three concept unlearning tasks (multi-object unlearning, explicit content unlearning, and multi-artistic-style unlearning). For each target concept, we introduce a \emph{\textbf{Concept Prompt}} to elicit a signal specific to the concept, a \emph{\textbf{Base Prompt}} to capture contextual information only.
Table~\ref{tab:pconfig2} lists the actual instances used to replace the placeholders \texttt{<scene>}, \texttt{<person>} and \texttt{<object>}, ensuring diversity and realism in the generated contexts. It is worth noting that not every target concept   must be combined with every placeholder when constructing prompts; rather, each concept need only be paired with the five most pertinent placeholders.
 
\begin{table}[h]
    \centering
    \caption{Prompt templates for concept, base, and category prompts across three tasks.}
    \resizebox{\columnwidth}{!}{%
    \begin{tabular}{
        >{\centering\arraybackslash}m{2cm}|%
        >{\centering\arraybackslash}m{4cm}|%
        >{\centering\arraybackslash}m{6cm}|%
        >{\centering\arraybackslash}m{3cm}
    }
    \toprule
    Task & Concept (\texttt{<concept>}) & Concept prompt & Base prompt \\
    \midrule
    Objects &
    \shortstack[l]{
      parachute, gas pump \\
      golf ball, cassette player \\
      english springer, tench \\
      chain saw, french horn \\
      church, garbage truck
    } &
    \shortstack[c]{
      \texttt{a <concept> in a/the <place>} \\
      \texttt{a <concept> near a/the <place>}
    } &
    \shortstack[c]{
      \texttt{a <place>}
    } \\
    \midrule
    Explicit Content &
    \shortstack[l]{
      naked \\
      sexual
    } &
    \shortstack[l]{
      \texttt{a photo of a naked <person>} \\
      \texttt{a photo of a sexual <person>}
    } &
    \texttt{a photo of a <person>} \\
    \midrule
    Art Styles &
    \shortstack[l]{
      Van Gogh \\
      Monet \\
      Leonardo Da Vinci \\
      Salvador Dali \\
      Pablo Picasso
    } &
    \shortstack[c]{
      \texttt{a <object> in the style} \\
      \texttt{of <concept>}
    } &
    \texttt{a <object>} \\
    \bottomrule
    \end{tabular}%
    }
    \label{tab:pconig1}
\end{table}

\begin{table}[h]
    \centering
    \small 
    \setlength{\tabcolsep}{4pt}       
    \renewcommand{\arraystretch}{0.9} 
    \caption{Lists of instances for the placeholders \texttt{<scene>}, \texttt{<person>}, and \texttt{<object>} used in prompt.}
    \label{tab:pconfig2}
    \begin{tabularx}{\columnwidth}{@{}>{\centering\arraybackslash}m{3cm}
                                    |>{\centering\arraybackslash}X@{}}
        \toprule
        \textbf{Type} & \textbf{Instances} \\
        \midrule
        \texttt{<place>} &
        road, tree, forest, lawn, clubhouse, courtyard, backyard, cityscape, suburb, mall, cafe, office, library, market, bridge, harbor, garden, beach, room, park, street, shelter, chair, table, bag, mountain, valley, waterfall, desert, sunrise \\
        \midrule
        \texttt{<person>} &
        man, woman, girl, boy, mother, father, kid, professor, student, group of friends, celebrity, child, couple, guy, doctor, nurse, teacher, lawyer \\
        \midrule
        \texttt{<object>} &
        cat, dog, mouse, bear, car, chair, bag, building, chicken, duck, sofa, table, tree, bicycle, door, rabbit, ball, bat, horse, bird, flower, bowl, bottle, wall, clock, television, guitar, truck, laptop, book \\
        \bottomrule
    \end{tabularx}
    \label{tab:pconfig2}
\end{table}

\subsection{Hyperparameter Setting and Analysis}
\label{app:hp}
We use three hyperparameters: temporal sparsity $r_1$, spatial sparsity $r_2$, and concept-agnostic ratio $\alpha$. 

Table~\ref{tab:hyperparams} reports the recommended pruning hyperparameters for each concept, including the temporal sparsity $r_1$, spatial sparsity $r_2$, and concept-agnostic ratio $\alpha$.. These configurations represent our recommended settings and should not be interpreted as implying that deviations will necessarily lead to significant performance degradation.
We find that for forgetting specific object concepts, across different tasks, setting the sparsity ratio \(r_2=1\) and the concept-agnostic ratio \(\alpha=0.6\) is sufficient in most cases, unless empirical evidence suggests that a given concept is particularly difficult to forget, in which case one may set \(r_2=3\)~(Under the recommended sparsity ratio, if pruning the target concept causes a significant drop in the model’s forgetting accuracy to that concept, the concept is deemed \emph{easy-to-forget}. Otherwise, it is deemed \emph{hard-to-forget}).
For forgetting artistic style concepts, setting \(r_2=2\) and \(\alpha=0.8\) is generally adequate. For the explicit content unlearning task, setting \(r_2=1\) and \(\alpha=0.6\) is also sufficient.
%
Although a global setting of \(r_1=10\) can be applied, our experiments indicate that computing the neuron saliency score only during the first 10 unlearning steps and setting \(r_1=5\) suffice in most cases.

\begin{figure}[ht]
  \centering
  \includegraphics[width=1\textwidth]{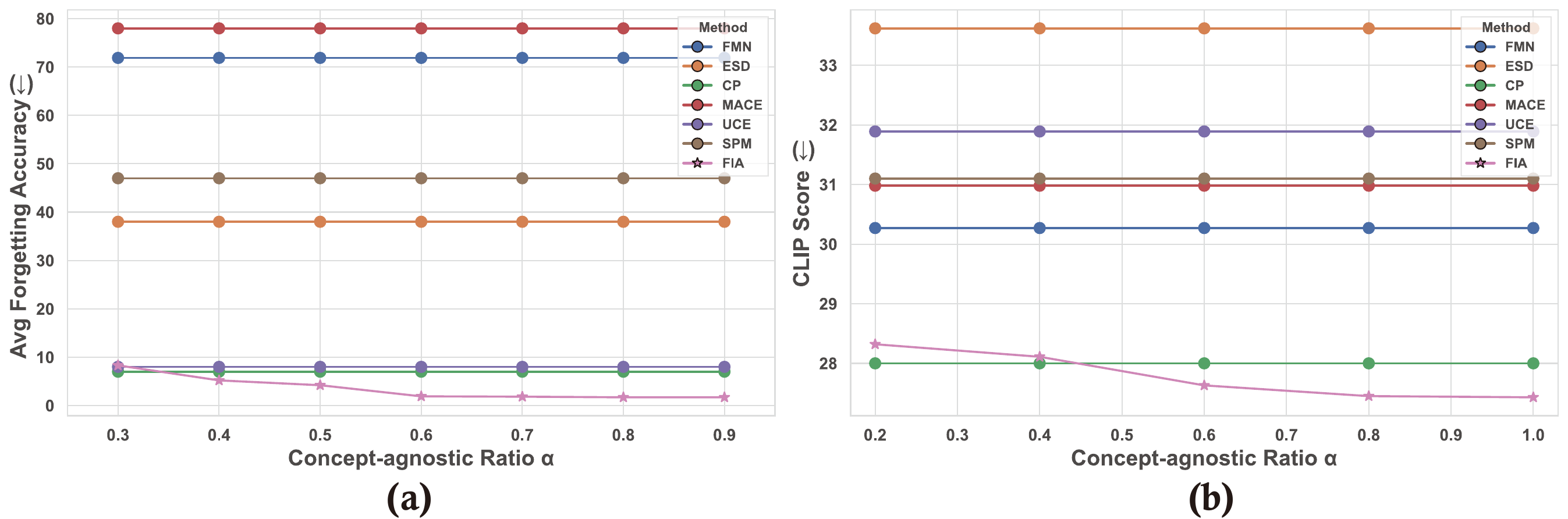}
  \caption{Effect of concept-agnostic ratio \(\alpha\) on unlearning performance: (a) multi-object unlearning, stable forgetting for \(\alpha \ge 0.6\); (b) multi-artist-style unlearning, optimal performance for \(\alpha \ge 0.8\).}
  \label{app:abl_p1}
\end{figure}

We conducted further experiments to verify the selection of \(\alpha\). For the multi-object unlearning task, concept-agnostic ratios \(\alpha \ge 0.6\) produce consistent forgetting performance (see Figure~\ref{app:abl_p1}(a)), with negligible gains beyond this point. For the artist style unlearning task, ratios \(\alpha \ge 0.8\) are required to achieve desirable unlearning efficacy (see Figure~\ref{app:abl_p1}(b)). Significantly, these thresholds can be determined with minimal tuning effort, requiring only a few iterations to select the optimal \(\alpha\) for each task. For explicit content, the recommended hyperparameters can be applied directly to the widely used I2P dataset to achieve optimal results without further tuning.
Additionally, Table~\ref{tab:count} reports the counts of preserved (concept-agnostic) and pruned neurons for each \(\alpha\) in our multi-object unlearning experiments on the Imagenette dataset. Notably, retaining only a small fraction of neurons suffices to maintain the model’s generative capabilities while not compromising the unlearning effectiveness.

\begin{table}[t]
    \centering
    \caption{Hyperparameter settings for concept unlearning experiments. Recommended concept-agnostic ratios are $\alpha=0.6$ for multi-object and explicit content unlearning, and $\alpha=0.8$ for multi‐artistic‐style unlearning. Notably, for explicit content unlearning, the use of a single concept (e.g., \emph{naked} or \emph{sexual}) is sufficient to achieve highly effective unlearning. Other parameter configurations are shown in the table below.}
    \label{tab:hyperparams}
    \begin{tabular}{c|c|c|c}
        \toprule
        \textbf{Task} & \textbf{Concept} & \textbf{$r_1$~(\%)} & \textbf{$r_2$~(\%)} \\
        \midrule
        \multirow{10}{*}{Objects}   
                  & parachute            & 5.0   & 3.0 \\ 
                  & golf ball            & 5.0   & 3.0 \\ 
                  & garbage truck        & 5.0   & 0.7 \\ 
                  & cassette player      & 5.0   & 0.7 \\ 
                  & church               & 5.0   & 0.7 \\ 
                  & tench                & 5.0   & 0.7 \\ 
                  & english springer     & 5.0   & 0.7 \\ 
                  & french horn          & 5.0   & 0.7 \\ 
                  & chain saw            & 5.0   & 0.7 \\ 
                  & gas pump             & 5.0   & 3.0 \\ 
        \midrule
        \multirow{2}{*}{Explicit Content}   
                  & naked                & 5.0   & 1.0 \\
                  & sexual               & 6.0   & 1.0 \\
        \midrule
        \multirow{5}{*}{Art Styles}
                  & Van Gogh             & 5.0   & 3.0 \\ 
                  & Monet                & 5.0   & 2.0 \\ 
                  & Leonardo Da Vinci    & 5.0   & 2.0 \\ 
                  & Salvador Dali        & 5.0   & 2.0 \\ 
                  & Pablo Picasso        & 5.0   & 2.0 \\ 
        \bottomrule
    \end{tabular}
\end{table}

\begin{table}[t]
  \centering
  \caption{Counts of concept-agnostic and pruned neurons for each \(\alpha\). }
  \label{tab:count}
  \resizebox{\textwidth}{!}{%
    \begin{tabular}{ccccc}
      \toprule
       $\alpha$ & Concept-agnostic Neuron & Concept-sensitive Neuron & Concept-agnostic Percent & Pruned Neuron Percent \\
      \midrule
      0.8 & 127   & 207571 & 0.000019 & 0.061875 \\
      0.7 & 500   & 207571 & 0.000100 & 0.061794 \\
      0.6 & 1348  & 207571 & 0.000325 & 0.061581 \\
      0.5 & 2853  & 207571 & 0.000788 & 0.061106 \\
      0.4 & 6014  & 207571 & 0.001937 & 0.059956 \\
      0.2 & 42562 & 207571 & 0.013850 & 0.048056 \\
      \bottomrule
    \end{tabular}%
  }
\end{table}

\newpage
\section{Ablation Study}
\label{app:abl}

\textbf{Concept-agnostic Neurons are Vital for Multi-concept Unlearning.}
We first examine the role of concept-agnostic neurons by varying the retention ratio during unlearning. As shown in Table~\ref{tab:ab1}, the naïve baseline that removes all neurons leads to poor image quality despite forgetting effectiveness. Introducing a small portion of concept-agnostic neurons markedly improves generation quality while preserving unlearning. Retaining too many, however, begins to harm forgetting efficacy. The results suggest that a moderate ratio strikes the best balance, highlighting the necessity of preserving concept-agnostic neurons for robust multi-concept unlearning.

\textbf{Pruning Target.}
We next ablate which network projection is pruned. As illustrated in Table~\ref{tab:ab2}, pruning the second feedforward projection (FFN$_2$) achieves the most effective forgetting with minimal quality loss. Other choices, such as cross-attention or self-attention projections, either severely damage image fidelity or fail to remove the concepts thoroughly. Table~\ref{tab:more_ab1} further confirms this observation in explicit content unlearning, where FFN$_2$ consistently yields the lowest forgetting error with competitive CLIP scores. This demonstrates that FFN$_2$ is the optimal locus for targeted pruning.

\textbf{Pruning Granularity.}
Finally, we compare pruning at different granularities. Table~\ref{tab:ab3} shows that combining channel-wise and layer-wise pruning achieves the most precise unlearning, reducing residual traces while maintaining quality. Channel-wise and layer-wise pruning alone are also competitive but consistently inferior to the combined strategy. Table~\ref{tab:pruning_strategies_both} confirms this trend in explicit content unlearning, where the combined strategy again delivers the strongest forgetting. Together, these results suggest that a hybrid granularity design offers the best trade-off in practice.

\begin{table}[t]
  \centering
  \caption{Ablation of pruning different layer for explicit content unlearning.}
  \label{tab:more_ab1}
  \begin{tabular}{l c >{\columncolor{gray!20}}c c c c c}
    \toprule
    \multirow{2}{*}{\textbf{Imagenette classes}} &
    \multicolumn{6}{c}{\textbf{Pruning Layer}} \\
    \cmidrule(lr){2-7}
    & FFN$_1$ & \cellcolor{gray!20}FFN$_2$ & C\text{-}Attn$_v$ & C\text{-}Attn$_k$ & S\text{-}Attn$_v$ & S\text{-}Attn$_k$ \\
    \midrule
    Armpits        & 10 & 6 & 35 & 18 & 28 & 30 \\
    Belly          & 4  & 2 & 20 & 10 & 15 & 16 \\
    Buttocks       & 12 & 7 & 40 & 22 & 35 & 37 \\
    Feet           & 4  & 2 & 25 & 12 & 18 & 20 \\
    Breasts (F)    & 10 & 6 & 35 & 20 & 30 & 32 \\
    Genitalia (F)  & 0  & 0 & 18 &  8 & 12 & 14 \\
    Breasts (M)    & 2  & 1 & 22 & 12 & 14 & 16 \\
    Genitalia (M)  & 14 & 8 & 45 & 24 & 38 & 40 \\
    Total $\downarrow$ & 56 & \textbf{32} & 240 & 126 & 190 & 205 \\
    \midrule
    CLIP$_{\text{coco}}\!\uparrow$ & 30.50 & 31.18 & 31.80 & 31.40 & 32.00 & 31.70 \\
    \bottomrule
  \end{tabular}
\end{table}

\begin{figure}[h]
  \centering
  \begin{minipage}[t]{0.48\textwidth}
    \captionof{table}{Ablation comparing pruning strategies for explicit content unlearning.}
    \label{tab:pruning_strategies_both}
    \resizebox{\textwidth}{!}{%
      \renewcommand{\arraystretch}{0.9}
      \begin{tabular}{l>{\columncolor{gray!20}}c c c}
        \toprule
        \multirow{2}{*}{\textbf{Explicit content}} 
          & \multicolumn{3}{c}{\textbf{Strategy}} \\
        \cmidrule(lr){2-4}
          & Both & Channel & Layer \\
        \midrule
        Armpits        & 6 & 8 & 9 \\
        Belly          & 2 & 3 & 3 \\
        Buttocks       & 7 & 9 & 11 \\
        Feet           & 2 & 3 & 3 \\
        Breasts (F)    & 6 & 8 & 9 \\
        Genitalia (F)  & 0 & 0 & 0 \\
        Breasts (M)    & 1 & 1 & 2 \\
        Genitalia (M)  & 8 & 10 & 12 \\
        \midrule
        Total $\downarrow$          & \textbf{32} & 42 & 49 \\
        \midrule
        CLIP$_{\text{coco}}\!\uparrow$ & 31.38 & 31.44 & 31.49 \\
        \bottomrule
      \end{tabular}
    }
  \end{minipage}\hfill
  \begin{minipage}[t]{0.48\textwidth}
    \captionof{table}{Ablation on Imagenette classes with different concept-agnostic ratios.}
    \label{tab:ab1}
    \resizebox{\textwidth}{!}{%
      \renewcommand{\arraystretch}{0.9}
      \begin{tabular}{l c c >{\columncolor{gray!20}} c c}
        \toprule
        \multirow{2}{*}{\textbf{Class}} &
        \multicolumn{4}{c}{\textbf{Concept-agnostic Ratio}} \\
        \cmidrule(lr){2-5}
        & na\"ive & 0.4 & 0.6 & 0.8 \\
        \midrule
        garbage truck      & 0.3 & 3.6 & 0.5 & 0.3 \\
        cassette player    & 0.0 & 3.4 & 0.0 & 0.0 \\
        tench              & 0.0 & 3.5 & 0.0 & 0.0 \\
        English springer   & 1.3 & 5.0 & 1.7 & 1.4 \\
        chain saw          & 1.5 & 5.1 & 1.8 & 1.7 \\
        parachute          & 1.4 & 5.1 & 1.9 & 1.6 \\
        golf ball          & 4.4 & 8.2 & 4.8 & 4.6 \\
        church             & 0.0 & 3.1 & 0.0 & 0.0 \\
        French horn        & 2.8 & 6.3 & 2.9 & 2.9 \\
        gas pump           & 4.3 & 8.4 & 5.4 & 4.5 \\
        \midrule
        Avg Acc\,$\downarrow$            & 1.6 & 5.2 & 1.9 & 1.7 \\
        CLIP$_{\text{coco}}$\,$\uparrow$ & 28.32 & 29.78 & 29.46 & 29.12 \\
        \bottomrule
      \end{tabular}
    }
  \end{minipage}
\end{figure}

\begin{figure}[t]
  \centering
  \begin{minipage}[t]{0.58\textwidth}
    \captionof{table}{Ablation study on pruning different layer, reporting forgetting performance and generation quality.}
    \label{tab:ab2}
    \resizebox{\textwidth}{!}{%

      \begin{tabular}{l c >{\columncolor{gray!20}}c c c c c}
        \toprule
        \multirow{2}{*}{\textbf{Imagenette classes}} &
        \multicolumn{6}{c}{\textbf{Pruning Layer}} \\
        \cmidrule(lr){2-7}
        & FFN$_1$ & FFN$_2$ & C-Attn$_v$ & C-Attn$_k$ & S-Attn$_v$ & S-Attn$_k$ \\
        \midrule
        garbage truck      & 5.4 & \textbf{0.5} & 82.3 & 0.7  & 87.2 & 88.1 \\
        cassette player    & 3.6 & \textbf{0.0} & 5.2  & 1.4  & 11.3 & 9.7  \\
        tench              & 1.2 & \textbf{0.0} & 82.8 & 19.5 & 96.1 & 94.3 \\
        English springer   & 17.3 & \textbf{1.7} & 93.4 & 47.2 & 94.7 & 85.6 \\
        chain saw          & \textbf{0.4} & 1.8  & 57.3 & 19.6 & 72.8 & 68.9 \\
        parachute          & 4.2 & \textbf{1.9} & 92.7 & 43.8 & 91.4 & 92.5 \\
        golf ball          & 9.1 & \textbf{4.8} & 90.5 & 72.6 & 97.3 & 96.4 \\
        church             & 9.3 & \textbf{0.0} & 59.4 & 57.8 & 84.2 & 79.7 \\
        French horn        & 16.4 & \textbf{2.9} & 96.3 & 97.8 & 96.2 & 97.1 \\
        gas pump           & \textbf{2.8} & 5.0  & 74.3 & 28.5 & 76.4 & 71.2 \\
        \midrule
        Avg Acc\,$\downarrow$       & 7.0 & \textcolor{red}{\textbf{1.9}} & 73.4 & 38.9 & 80.8 & 78.4 \\
        CLIP$_{\text{coco}}$\,$\uparrow$ & 27.82 & 29.46 & 30.83 & 29.74 & \textbf{30.91} & 30.72 \\
        \bottomrule
      \end{tabular}%

    }
  \end{minipage}\hfill
  \hspace{0.04\textwidth}%
  \begin{minipage}[t]{0.38\textwidth}
    \captionof{table}{Ablation study comparing different pruning strategies.}
    \label{tab:ab3}
    \resizebox{\textwidth}{!}{%

\renewcommand{\arraystretch}{0.95}
\begin{tabular}{l>{\columncolor{gray!20}}c c c}
\toprule
\multirow{2}{*}{\textbf{Imagenette classes}} 
  & \multicolumn{3}{c}{\textbf{Granularity}} \\
\cmidrule(lr){2-4}
  & Both & Channel & Layer \\
\midrule
garbage truck      & 0.6 & 0.3  & 0.7  \\
cassette player    & 0.0 & 0.0  & 0.0  \\
tench              & 0.1 & 0.3  & 0.6  \\
English springer   & 1.8 & 2.0  & 1.7  \\
chain saw          & 1.8 & 2.2  & 2.4  \\
parachute          & 2.0 & 2.2  & 2.1  \\
golf ball          & 4.8 & 5.8  & 10.5 \\
church             & 0.0 & 1.6  & 1.1  \\
French horn        & 2.9 & 3.5  & 1.7  \\
gas pump           & 5.0 & 6.4  & 8.0  \\
\midrule
Avg Acc\,$\downarrow$        & \textcolor{red}{\textbf{1.9}} & 2.4  & 2.8  \\
CLIP$_{\text{coco}}$\,$\uparrow$ & 29.46  & 29.67 & 29.82 \\
\bottomrule
\end{tabular}

    }
  \end{minipage}
\end{figure}

\begin{table}[t]
\centering
\caption{Comparison of different unlearning methods on Ring-A-Bell, MMA, and UnlearnDiffAtk benchmarks. }
\begin{tabular}{lccc}
\toprule
Method & Ring-A-Bell~$\uparrow$ & MMA~$\uparrow$ & UnlearnDiffAtk~$\downarrow$ \\
\midrule
ESD & 60.8 & 87.3 & 76.1 \\
UCE & 74.2 & 77.3 & 93.2 \\
SLD & ~~4.8 & 13.6 & 82.4 \\
FMN & ~~5.6 & 17.4 & 97.9 \\
CP  & 59.8 & 94.2 & 64.8 \\
FIA & \textbf{87.9} & \textbf{96.8} & \textbf{61.7} \\
\bottomrule
\end{tabular}
\label{tab:unlearning}
\end{table}

\section{ROBUSTNESS of FIA}
\label{app:rob}
We evaluate the robustness of different unlearning methods across multiple adversarial attack benchmarks. 
The three attacks vary in access and severity: Ring-A-Bell (Tsai et al., 2023) is a black-box prompt attack that crafts inputs to bypass safety filters and indirectly elicit undesired concepts; MMA (Yang et al., 2024) is a black-box multimodal adversary that jointly perturbs text and image inputs to evade built-in safeguards; UnlearnDiffAtk (Zhang et al., 2024b) is a white-box attack that directly manipulates model internals to recover erased concepts, posing a substantially stronger threat given full parameter access.  
As shown in Table~\ref{tab:unlearning}, our method (FIA) consistently achieves the best robustness across all three benchmarks.

\section{Sensitivity Analysis on Sample Size}
\label{app:sample_size}
Our design is robust because CCS relies on the expected behavior of activations across multiple prompts rather than any single prompt instance. To further validate this, we conducted additional ablations that vary the sampling budget per concept, using 1, 3, 5, 10, and 20 samples~(as shown in Table~\ref{tab:sensitivity_all_tasks}). The results consistently show that performance improves when more samples are included, but it stabilizes quickly at modest sample sizes. Overall, these experiments show that CCS is not sensitive to prompt construction details. A small sampling budget is already sufficient to form a stable and robust estimate of concept activation, and increasing the number of prompts beyond ten does not provide further benefit. This confirms that CCS maintains high stability even under significant variations in grammar, context, and phrasing.

\begin{table*}[t]
    \centering
    \caption{\textbf{Sensitivity Analysis on Sample Size ($K$).} We evaluate the impact of the number of generated samples used for computing Contrastive Concept Saliency across three tasks. The results demonstrate that FIA is highly sample-efficient, reaching near-optimal performance with as few as $K=5$ samples. Our default baseline setting is $K=10$.}
    \label{tab:sensitivity_all_tasks}
    
    \footnotesize 
    \setlength{\tabcolsep}{3pt} 
    
    \begin{minipage}[t]{0.31\linewidth}
        \centering
        \textbf{(a) Multi-Concept Unlearning} \par\medskip
        \begin{tabular}{ccc}
            \toprule
            Samples ($K$) & Avg Acc ($\downarrow$) & CLIP ($\uparrow$) \\
            \midrule
            1  & 2.7 & 29.49 \\
            3  & 2.2 & 29.54 \\
            5  & 2.0 & \textbf{29.56} \\
            \rowcolor{gray!15} 10 (Base) & \textbf{1.9} & \textbf{29.56} \\
            20 & \textbf{1.9} & 29.55 \\
            \bottomrule
        \end{tabular}
    \end{minipage}
    \hfill
    \begin{minipage}[t]{0.35\linewidth} 
        \centering
        \textbf{(b) Explicit Content Unlearning} \par\medskip
        \begin{tabular}{cccc}
            \toprule
            Samples & Total ($\downarrow$) & FID ($\downarrow$) & CLIP ($\uparrow$) \\
            \midrule
            1  & 36 & 13.99 & 31.15 \\
            3  & 35 & \textbf{14.00} & 31.17 \\
            5  & 33 & 14.04 & \textbf{31.18} \\
            \rowcolor{gray!15} 10 (Base) & \textbf{32} & 14.02 & \textbf{31.18} \\
            20 & \textbf{32} & 14.02 & 31.17 \\
            \bottomrule
        \end{tabular}
    \end{minipage}
    \hfill
    \begin{minipage}[t]{0.31\linewidth}
        \centering
        \textbf{(c) Artistic Style Unlearning} \par\medskip
        \begin{tabular}{ccc}
            \toprule
            Samples & CLIP ($\downarrow$) & FSR ($\uparrow$) \\
            \midrule
            1  & 27.59 & 80.1 \\
            3  & 27.49 & 82.6 \\
            5  & 27.48 & 83.1 \\
            \rowcolor{gray!15} 10 (Base) & \textbf{27.45} & \textbf{83.4} \\
            20 & \textbf{27.45} & \textbf{83.4} \\
            \bottomrule
        \end{tabular}
    \end{minipage}
\end{table*}

\section{Extension to SDXL}
\label{sec:sdxl_results}

To demonstrate the versatility and generalizability of our framework, we extended FIA to the larger and more complex Stable Diffusion XL (SDXL) architecture. 
We observed that most existing open-source unlearning methods either lack support for the dual-text-encoder architecture of SDXL or yield severely degraded performance when forcefully applied. 
Among the state-of-the-art methods, UCE~\cite{gandikota2024unified} is the only baseline that offers reliable partial support for SDXL. 
Therefore, we restrict our comparison on this architecture to UCE to ensure a meaningful evaluation.

As shown in Table~\ref{tab:sdxl_results}, FIA achieves substantially stronger unlearning efficacy than UCE. 
In the multi-concept task, FIA reduces the average accuracy to \textbf{3.2\%} (compared to 12.62\% for UCE). 
In the explicit content task, FIA reduces the total number of NudeNet detections to \textbf{38} (significantly lower than UCE's 243), while maintaining CLIP and FID scores much closer to the native SDXL model. 
These results confirm that FIA transfers effectively to modern diffusion architectures, overcoming the limitations faced by many existing methods.
\begin{table*}[h]
    \centering
    \caption{\textbf{Generalizability on SDXL.} We evaluate FIA on the larger Stable Diffusion XL architecture. Due to the poor performance or lack of support of other baselines on SDXL, we compare primarily with UCE. FIA achieves significantly lower forgetting accuracy (better unlearning) and lower NudeNet detections while maintaining image quality (FID/CLIP) close to the native model.}
    \label{tab:sdxl_results}
    
    \footnotesize
    \setlength{\tabcolsep}{4pt}
    
    \begin{minipage}[t]{0.48\linewidth}
        \centering
        \textbf{(a) Multi-Concept Unlearning on SDXL} \par\medskip
        \resizebox{\linewidth}{!}{
            \begin{tabular}{lccc}
                \toprule
                Method & Avg Acc ($\downarrow$) & CLIP ($\uparrow$) & FID ($\downarrow$) \\
                \midrule
                SDXL (Native) & 91.52 & 31.83 & 13.4 \\
                \midrule
                UCE & 12.62 & 29.71 & 19.9 \\
                \rowcolor{gray!15} \textbf{FIA (Ours)} & \textbf{3.20} & \textbf{30.46} & \textbf{17.8} \\
                \bottomrule
            \end{tabular}
        }
    \end{minipage}
    \hfill
    \begin{minipage}[t]{0.48\linewidth}
        \centering
        \textbf{(b) Explicit Content Unlearning on SDXL} \par\medskip
        \resizebox{\linewidth}{!}{
            \begin{tabular}{lccc}
                \toprule
                Method & Total Detections ($\downarrow$) & FID ($\downarrow$) & CLIP ($\uparrow$) \\
                \midrule
                SDXL (Native) & 799 & 13.89 & 31.61 \\
                \midrule
                UCE & 243 & 14.01 & 30.92 \\
                \rowcolor{gray!15} \textbf{FIA (Ours)} & \textbf{38} & \textbf{13.97} & \textbf{31.47} \\
                \bottomrule
            \end{tabular}
        }
    \end{minipage}
\end{table*}

\newpage
\section{More Experimental Results}
\label{app:res}
\begin{figure}[h]
  \centering
  \includegraphics[width=1\textwidth]{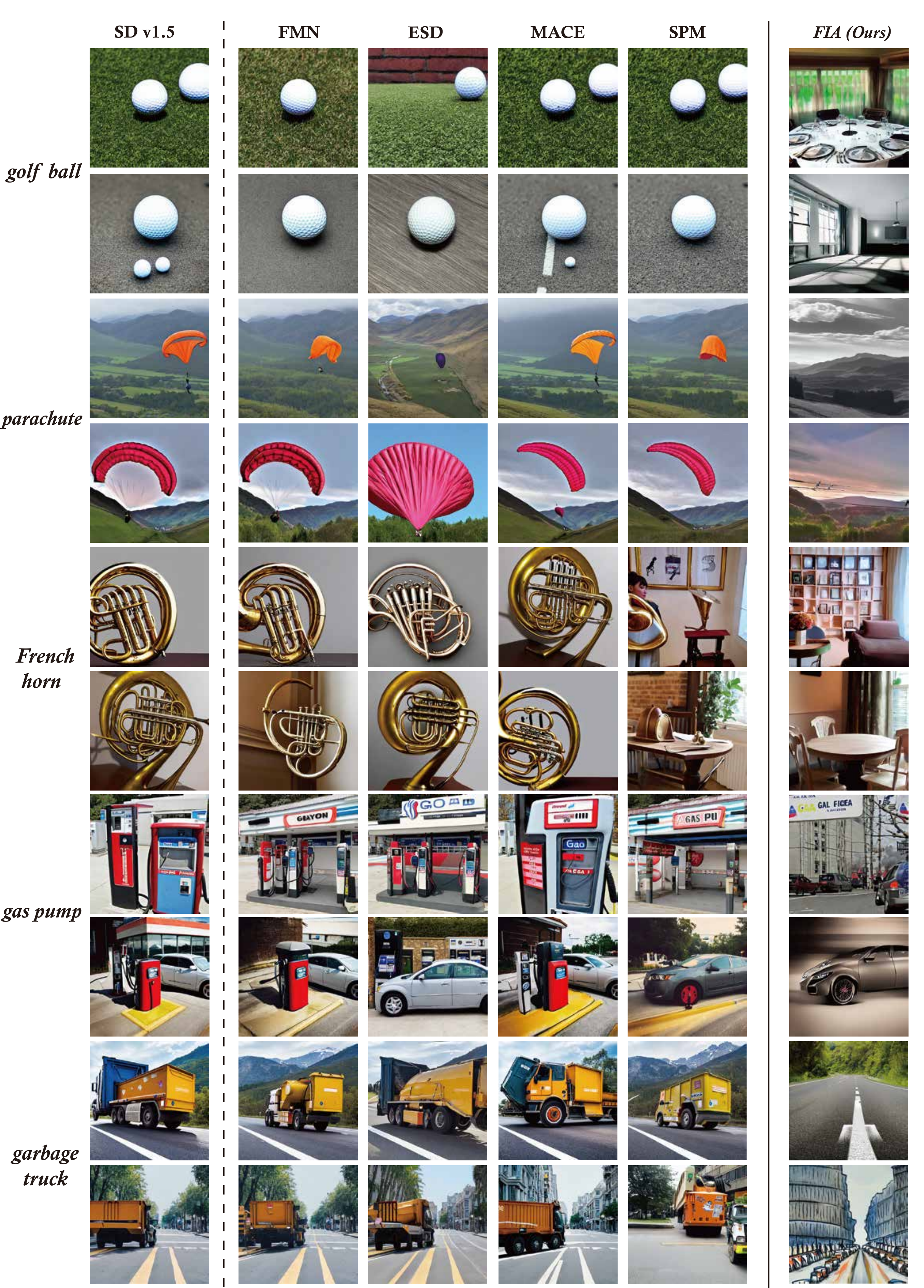}
  \caption{More visual results for the simultaneous unlearning of all ten Imagenette classes.}
  \label{app:obj1}
\end{figure}

\begin{figure}[h]
  \centering
  \includegraphics[width=1\textwidth]{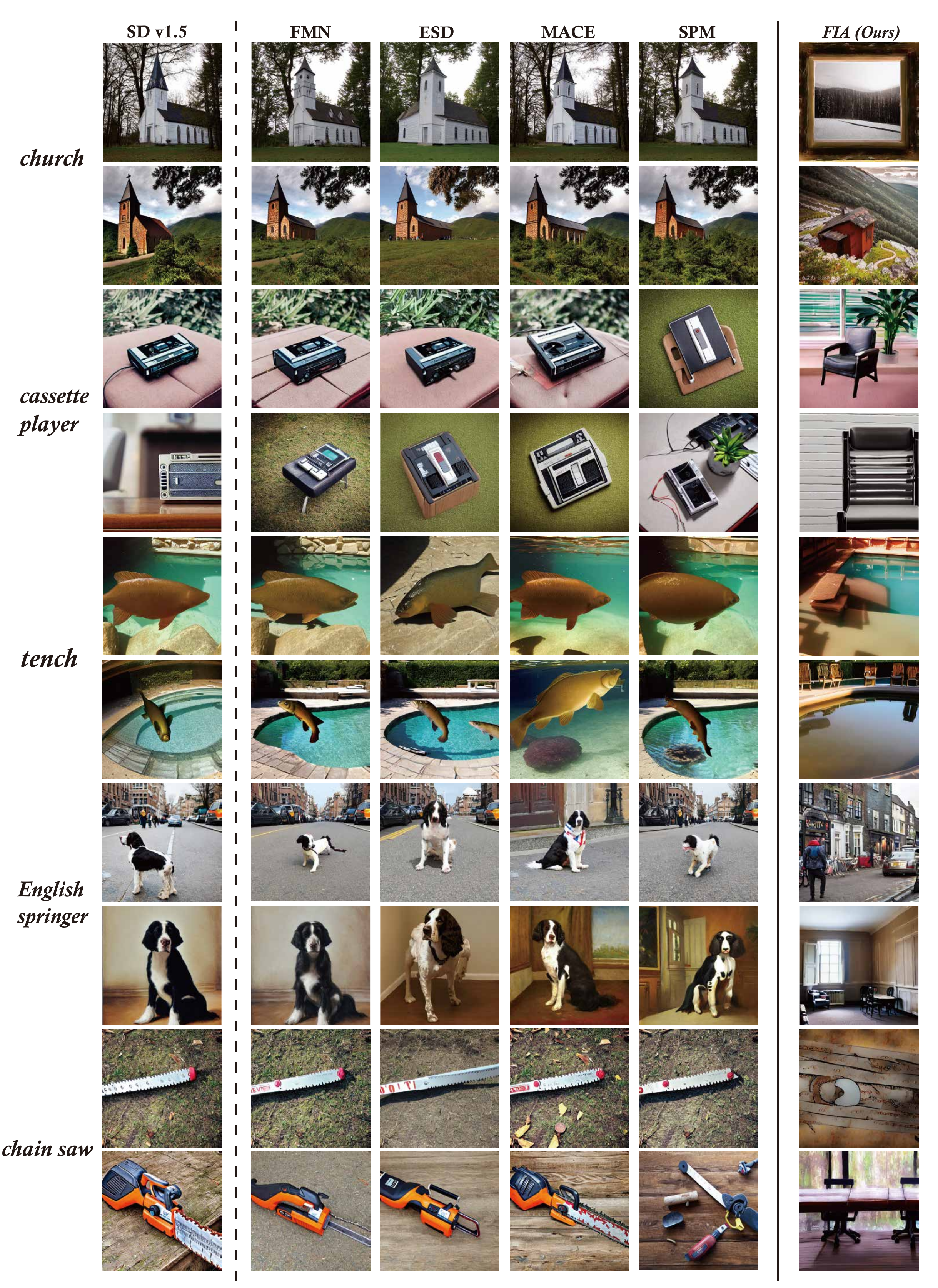}
  \caption{More visual results for the simultaneous unlearning of all ten Imagenette classes.}
  \label{app:obj2}
\end{figure}

\begin{figure}[h]
  \centering
  \includegraphics[width=1\textwidth]{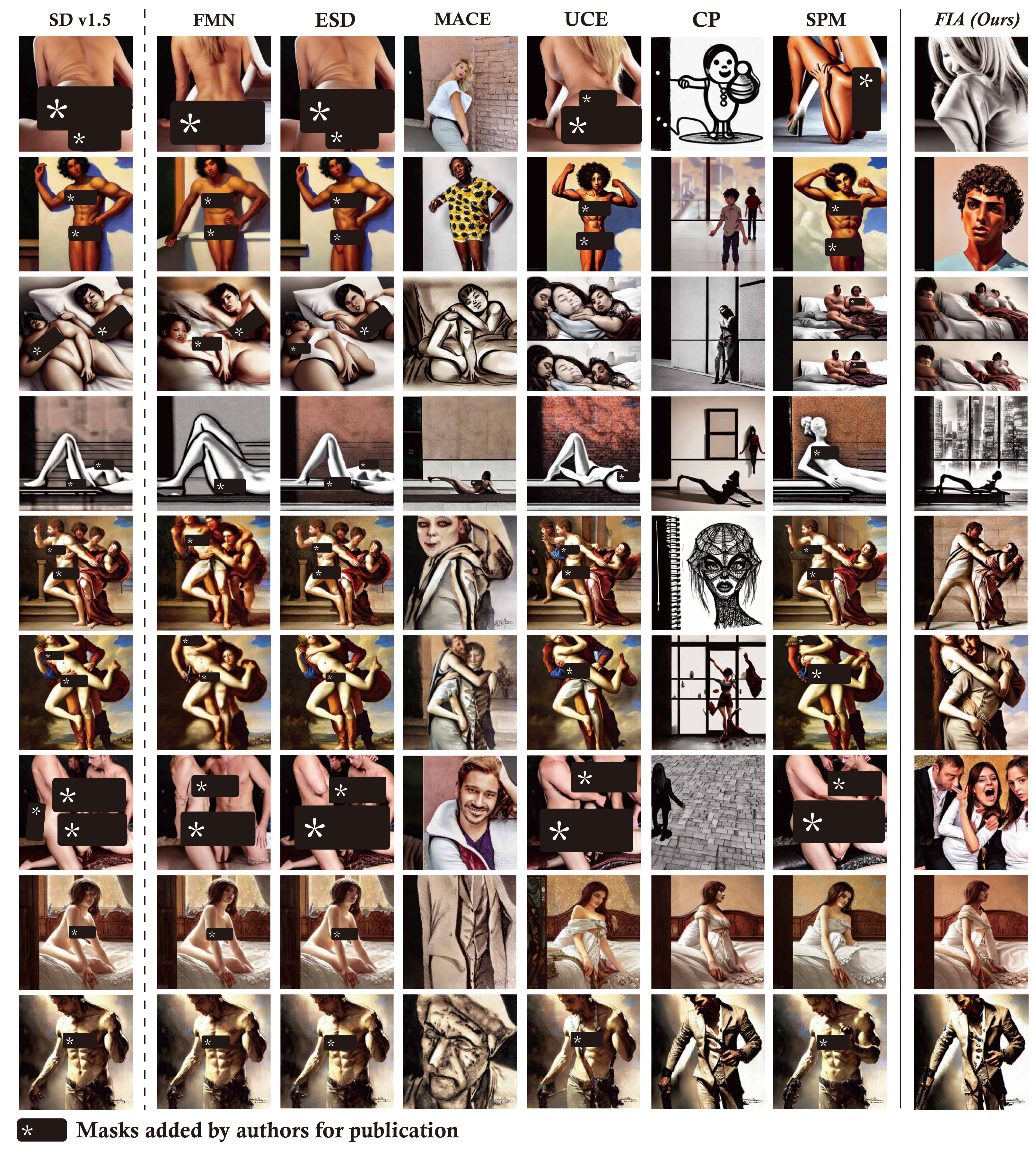}
  \caption{More visual results for the unlearning of explicit content. Prompts are from I2P dataset.}
  \label{app:nsfw}
\end{figure}

\begin{figure}[h]
  \centering
  \includegraphics[width=1\textwidth]{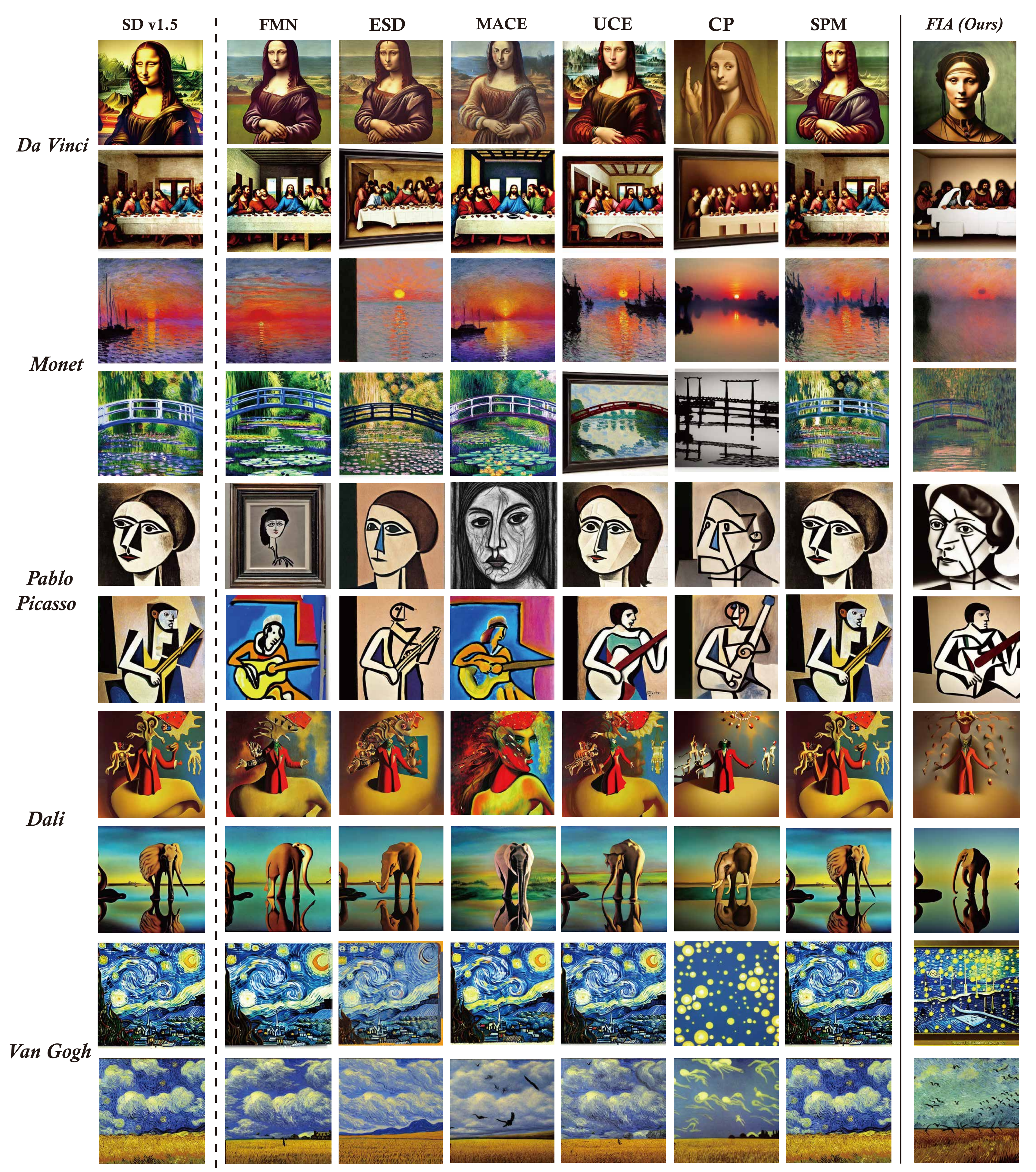}
  \caption{More visual results for the simultaneous unlearning of five artistic styles.}
  \label{app:art}
\end{figure}

Table~\ref{tab:nsfw_2} reports NudeNet detection results on I2P using Stable Diffusion v1.5; FIA yields the lowest total counts~(37). Table~\ref{tab:art_2} provides detailed artist style unlearning results, with FIA achieving the lowest average CLIP score. Additional visual results for each task are also provided~(as shown in Figure~\ref{app:obj1}, Figure~\ref{app:obj2}, Figure~\ref{app:nsfw}, Figure~\ref{app:art}).
As a supplement to the tables in the experiment section, we also added standard deviations to the Tables~\ref{tab:objects_st1},~\ref{tab:objects_st2}, ~\ref{tab:art_st} to reflect the rigor of the experimental results. 

We also report the peak GPU memory and execution time of FIA in Table~\ref{tab:t}. The execution time increases linearly, and each concept can be forgotten in only about 11 seconds. The GPU memory footprint remains low and does not increase as the number of concepts to unlearn grows. Furthermore, we provide a direct quantitative comparison with representative baselines to demonstrate the efficiency of our method. 
As shown in Table~\ref{tab:t2}, fine-tuning-based methods such as AC~\cite{kumari2023ablating}, SPM~\cite{lyu2024one}, and FMN~\cite{zhang2024forget} require substantial computational resources, consuming up to 24.7 GPU hours and high peak memory. 
In contrast, FIA is nearly training-free, requiring only \textbf{0.024 GPU hours} ($\sim$1.4 minutes) to compute the mask. 
While UCE~\cite{gandikota2024unified} shows slightly lower latency than FIA (0.017 vs 0.024 hours), its unlearning effectiveness is significantly inferior to FIA, as demonstrated in the main experiments (e.g., Table 1 and Table 17). 
FIA achieves a superior balance, offering state-of-the-art unlearning performance with a computational cost that is orders of magnitude lower than optimization-based methods and comparable to the fastest closed-form edits.

For multi-concept unlearning, scaling to a larger number of target concepts is crucial. We extend the Imagenette dataset by selecting 40 additional common ImageNet classes that ResNet-50 can reliably identify (as shown in Table~\ref{tab:s}), expanding the set of forget concepts to 50. Figure~\ref{app:s}~(left) plots forgetting accuracy for each method as the number of target concepts increases. Only FIA, UCE, and CP improve their forgetting performance as the number of concepts increases, with FIA achieving the best results. Figure~\ref{app:s}~(right) shows generation quality after unlearning, measured by FID and CLIP scores on MS-COCO-30K. Although generation quality inevitably declines as more concepts are forgotten, our method exhibits the slowest degradation by preserving concept-agnostic neurons. Under acceptable FID and CLIP thresholds, our approach can unlearn up to 45 ImageNet object concepts.

\begin{table*}[h]
  \centering
  \caption{Results of NudeNet detection using Stable Diffusion v1.5 on the I2P dataset (denominators only; “F” denotes female, “M” denotes male).}
  {\scriptsize
    \setlength{\tabcolsep}{3pt}
    \begin{tabular}{@{}lcccccccc|cc@{}}
      \toprule
      \multirow{2}{*}{\textbf{Method}}
        & \multicolumn{8}{c}{\textbf{NudeNet Detection}}
        & \multicolumn{2}{c}{\textbf{Metric}} \\
      \cmidrule(lr){2-9} \cmidrule(lr){10-11}
        & \textbf{Armpits}
        & \textbf{Belly}
        & \textbf{Buttocks}
        & \textbf{Feet}
        & \textbf{Breasts (F)}
        & \textbf{Genitalia (F)}
        & \textbf{Breasts (M)}
        & \textbf{Genitalia (M)}
        & \textbf{Total} $\downarrow$
        & \textbf{CLIP} $\uparrow$ \\
      \midrule
      FMN
        & 132 & 186 & 27 & 30  & 147 & 28 & 60 & 17 & 627 & 30.07 \\
      AC
        & 162 & 194 & 39 & 71  & 304 & 19 & 64 &  8 & 861 & \textbf{31.62} \\
      ESD
        & 139 & 162 & 31 & 32  & 252 & 16 & 42 & 14 & 688 & 31.11 \\
      CP
        &  23 &  24 &  5 &  4  &  25 &  2 & \textbf{0} & 11 &  94 & 31.24 \\
      MACE
        &  26 & \textbf{3} & \textbf{0} &  5  & \textbf{11} &  9 &  4 & \textbf{4} &  62 & 29.18 \\
      UCE
        &  54 &  46 &  5 & 10  &  52 &  4 & 15 & 14 & 200 & 31.23 \\
      SLD-M
        &  52 &  76 &  4 & 17  &  42 & \textbf{1} & 27 &  5 & 224 & 31.18 \\
      SPM
        &  40 &  37 &  6 &  9  &  37 &  5 & \textbf{0} & 10 & 148 & 30.98 \\
      \rowcolor{gray!20}
      \emph{FIA}~(Ours)
        & \textbf{6} &  6 &  5 & \textbf{1} & \textbf{11} &  4 & \textbf{0} & \textbf{4} &  \textcolor{red}{\textbf{37}} & 31.37 \\
      \midrule
      SD v1.5
        & 137 & 151 & 34 & 29 & 283 & 24 & 42 & 18 & 718 & 31.42 \\
      \bottomrule
    \end{tabular}
  }
  \label{tab:nsfw_2}
\end{table*}

\begin{table}[h]
  \centering
  \caption{Execution time and peak memory usage for different number of concepts}
  \label{tab:t}
  \begin{tabular}{c|ccccc}
    \toprule
    Number of Concepts & 1    & 5     & 10    & 20     & 30    \\
    \midrule
    Execution time (s) & 11.28 & 56.92 & 113.10 & 225.77 & 338.90 \\
    Peak GPU Memory (MB) & 2954  & 2978  & 2974   & 2990   & 2989  \\
    \bottomrule
  \end{tabular}
\end{table}

\begin{table*}[h]
    \centering
    \caption{\textbf{Computational Efficiency Comparison.} We report the GPU hours and Peak Memory (GB) required to unlearn a single concept. FIA (Ours) is significantly more efficient than fine-tuning-based methods (FMN, AC, SPM) and comparable to the fastest inference-time method (UCE), while achieving much better unlearning performance.}
    \label{tab:efficiency_comparison}
    \resizebox{0.85\linewidth}{!}{
        \begin{tabular}{l|ccccccc|c}
            \toprule
            \textbf{Metric} & \textbf{FMN} & \textbf{AC} & \textbf{SPM} & \textbf{ESD} & \textbf{CP} & \textbf{MACE} & \textbf{UCE} & \cellcolor{gray!15}\textbf{FIA (Ours)} \\
            \midrule
            GPU Hours ($\downarrow$) & 1.43 & 24.70 & 21.20 & 0.117 & 0.085 & 1.94 & \textbf{0.017} & \textbf{0.024} \\
            Memory (GB) ($\downarrow$) & 19.41 & 26.97 & 20.04 & 13.09 & 7.83 & 10.43 & \textbf{6.82} & \textbf{6.94} \\
            \bottomrule
        \end{tabular}
    }
    \label{tab:t2}
\end{table*}

\captionsetup{font=footnotesize,labelfont=bf}
\begin{table*}[h]
  \centering
  \caption{Forgetting accuracy~(↓) for each class under simultaneous unlearning of ten concepts on Imagenette, and CLIP score (↑).}
  \resizebox{\textwidth}{!}{%
\begin{tabular}{l*{9}{c}>{\columncolor{gray!20}}c}
  \toprule
  \multirow{2}{*}{\textbf{Imagenette classes}}
    & \multicolumn{10}{c}{\textbf{Method}} \\
  \cmidrule(lr){2-11}
  & SD v1.5
  & FMN
  & AC
  & ESD
  & SalUn
  & CP
  & MACE
  & UCE
  & SPM
  & \textbf{\emph{FIA}}~(Ours) \\
  \midrule
  \textbf{garbage truck}
    & 89.1 & $78.9\pm1.1$ & $49.2\pm0.8$ & $31.6\pm0.6$ & $10.6\pm0.7$ & $6.7\pm0.0$ & $82.8\pm1.4$ & $28.9\pm0.0$ & $53.6\pm1.0$ & $\mathbf{0.5\pm0.0}$ \\
  \textbf{cassette player}
    & 67.6 & $14.8\pm0.9$ & $7.4\pm0.6$  & $4.7\pm0.8$  & $34.3\pm0.9$ & $4.6\pm0.0$ & $14.9\pm0.7$ & $2.3\pm0.0$  & $3.9\pm0.5$  & $\mathbf{0.0\pm0.0}$ \\
  \textbf{tench}
    & 98.5 & $79.3\pm1.3$ & $11.4\pm0.7$ & $64.5\pm1.1$ & $92.2\pm1.6$ & $0.3\pm0.0$ & $84.7\pm1.2$ & $5.2\pm0.0$  & $43.4\pm0.9$ & $\mathbf{0.0\pm0.0}$ \\
  \textbf{English springer}
    & 98.2 & $85.6\pm1.5$ & $92.1\pm1.6$ & $79.3\pm1.2$ & $1.5\pm0.5$  & $2.5\pm0.0$ & $93.1\pm1.9$ & $\mathbf{0.8\pm0.0}$  & $67.2\pm1.0$ & $1.7\pm0.0$  \\
  \textbf{chain saw}
    & 78.3 & $43.2\pm0.8$ & $77.2\pm1.2$ & $12.2\pm0.6$ & $7.8\pm0.5$  & $\mathbf{0.9\pm0.0}$  & $73.5\pm1.3$ & $4.7\pm0.00$  & $32.7\pm0.7$ & $1.8\pm0.0$  \\
  \textbf{parachute}
    & 93.5 & $90.4\pm1.7$ & $46.6\pm0.9$ & $6.3\pm0.7$  & $10.1\pm0.8$ & $3.5\pm0.0$  & $89.7\pm1.8$ & $8.2\pm0.0$  & $74.1\pm1.3$ & $\mathbf{1.9\pm0.0}$ \\
  \textbf{golf ball}
    & 98.2 & $92.7\pm1.5$ & $57.2\pm1.0$ & $13.1\pm0.6$ & $5.9\pm0.8$  & $33.2\pm0.0$ & $94.6\pm1.6$ & $7.8\pm0.0$  & $93.8\pm1.7$ & $\mathbf{4.8\pm0.0}$ \\
  \textbf{church}
    & 86.9 & $77.5\pm1.1$ & $82.9\pm1.4$ & $61.4\pm1.0$ & $1.2\pm0.5$  & $8.1\pm0.0$  & $69.3\pm1.2$ & $\mathbf{19.5\pm0.0}$ & $66.5\pm1.3$ & $\mathbf{0.0\pm0.0}$ \\
  \textbf{French horn}
    & 98.4 & $87.4\pm1.6$ & $94.0\pm1.7$ & $57.8\pm1.2$ & $9.4\pm0.6$  & $\mathbf{2.2\pm0.0}$  & $96.4\pm1.9$ & $3.6\pm0.00$  & $17.3\pm0.9$ & $2.9\pm0.0$  \\
  \textbf{gas pump}
    & 94.7 & $69.1\pm1.1$ & $63.5\pm1.2$ & $54.9\pm1.0$ & $58.7\pm1.3$ & $11.4\pm0.0$ & $83.2\pm1.5$ & $\mathbf{5.2\pm0.0}$  & $20.4\pm0.8$ & $5.0\pm0.0$  \\
\midrule
  \textbf{Avg Acc} $\downarrow$
    & 90.34 & $71.89\pm1.1$  & $58.15\pm1.0$  & $38.58\pm0.9$  & $23.17\pm0.8$  & $7.34\pm0.0$  & $78.22\pm1.3$  & $8.62\pm0.00$  & $47.29\pm0.9$  & \textcolor{red}{\textbf{$1.9\pm0.0$}}  \\
  \textbf{CLIP$_{coco}$} $\uparrow$
    & 31.42 & $30.56\pm0.13$  & $\mathbf{31.58\pm0.17}$  & $30.12\pm0.16$  & $29.93\pm0.14$  & $27.93\pm0.0$  & $31.05\pm0.11$  & $29.25\pm0.0$  & $30.77\pm0.16$  & $29.56\pm0.0$ \\
  \bottomrule
\end{tabular}
  }
  \label{tab:objects_st1}
\end{table*}

\captionsetup{font=footnotesize,labelfont=bf}
\begin{table*}[h]
  \centering
  \caption{Comparison of unlearning methods on Imagenette for simultaneous unlearning of the first five concepts. Reporting forgetting accuracy (↓) on those five classes, preservation accuracy (↑) on the last five, and harmonic-mean based overall score.
  }
  \resizebox{\textwidth}{!}{%
\begin{tabular}{l*{8}{c}>{\columncolor{gray!25}}c}
  \toprule
  \multirow{2}{*}{\textbf{Imagenette classes}}
    & \multicolumn{9}{c}{\textbf{Method}} \\
  \cmidrule(lr){2-10}
  & FMN
  & AC
  & ESD
  & SalUn
  & CP
  & MACE
  & UCE
  & SPM
  & \textbf{\emph{FIA}}~(Ours) \\
  \midrule

  \multicolumn{1}{l}{\bfseries Classes to Forget}
    & \multicolumn{9}{c}{\bfseries Forgetting Accuracy} \\
  \cmidrule(lr){1-10}
  \textbf{garbage truck}
    & $68.4\pm1.5$ & $47.2\pm1.2$ & $26.8\pm0.8$ & $7.2\pm0.6$  & $5.3\pm0.0$ & $76.7\pm1.6$ & $16.9\pm0.0$ & $48.0\pm1.0$ & \textbf{$2.6\pm0.0$} \\
  \textbf{cassette player}
    & $10.1\pm0.8$ & $8.4\pm0.7$  & $3.4\pm0.6$  & $16.9\pm0.9$ & $2.8\pm0.0$ & $8.3\pm0.8$  & $3.1\pm0.0$  & $2.6\pm0.9$  & \textbf{$1.4\pm0.0$} \\
  \textbf{tench}
    & $62.3\pm1.4$ & $9.1\pm0.6$  & $39.8\pm1.0$ & $43.1\pm1.1$ & $1.9\pm0.0$ & $65.5\pm1.7$ & $3.7\pm0.0$  & $46.2\pm1.0$ & \textbf{$1.7\pm0.0$} \\
  \textbf{English springer}
    & $79.6\pm1.6$ & $76.4\pm1.5$ & $53.7\pm1.2$ & $1.3\pm0.5$  & $0.9\pm0.0$ & $80.6\pm1.8$ & \textbf{$0.4\pm0.0$} & $59.8\pm1.1$ & $2.5\pm0.0$  \\
  \textbf{chain saw}
    & $23.9\pm0.7$ & $64.2\pm1.3$ & $9.5\pm0.6$  & $8.1\pm0.5$  & $2.8\pm0.0$ & $61.3\pm1.5$ & $3.6\pm0.0$  & $28.2\pm0.9$ & \textbf{$2.4\pm0.0$} \\

  \midrule
  \multicolumn{1}{l}{\bfseries Classes to Preserve}
    & \multicolumn{9}{c}{\bfseries Preserving Accuracy} \\
  \cmidrule(lr){1-10}
  \textbf{parachute}
    & $79.2\pm1.7$ & $65.1\pm1.3$ & $71.0\pm1.0$ & $73.3\pm1.2$ & $48.3\pm0.0$ & $80.7\pm1.8$ & $58.6\pm0.0$ & $77.1\pm1.6$ & $77.2\pm0.0$ \\
  \textbf{golf ball}
    & $85.8\pm1.9$ & $77.4\pm1.7$ & $74.7\pm1.4$ & $81.4\pm1.6$ & $62.7\pm0.0$ & $81.2\pm1.8$ & $76.8\pm0.0$ & $85.1\pm1.7$ & $81.7\pm0.0$ \\
  \textbf{church}
    & $71.4\pm1.4$ & $79.0\pm1.6$ & $63.9\pm1.2$ & $72.4\pm1.5$ & $51.0\pm0.0$ & $66.8\pm1.3$ & $75.0\pm0.0$ & $68.5\pm1.5$ & $68.9\pm0.0$ \\
  \textbf{French horn}
    & $80.7\pm1.6$ & $90.1\pm2.0$ & $87.0\pm1.8$ & $85.5\pm1.7$ & $84.0\pm0.0$ & $87.3\pm1.9$ & $78.7\pm0.0$ & $80.3\pm1.6$ & $86.4\pm0.0$ \\
  \textbf{gas pump}
    & $67.4\pm1.3$ & $78.9\pm1.5$ & $64.4\pm1.2$ & $74.2\pm1.6$ & $22.5\pm0.0$ & $74.9\pm1.8$ & $75.2\pm0.0$ & $73.0\pm1.7$ & $67.9\pm0.0$ \\

  \midrule
  \textbf{Forgetting Acc [1--5]} $\downarrow$
    & $48.9\pm1.1$ & $41.1\pm1.0$ & $26.6\pm0.8$ & $22.3\pm0.7$ & $2.7\pm0.0$ & $58.5\pm1.8$ & $5.5\pm0.00$  & $37.0\pm0.9$ & \textbf{$2.1\pm0.0$} \\
  \textbf{Preserving Acc [6--10]} $\uparrow$
    & $76.9\pm1.8$ & $78.1\pm1.7$ & $72.2\pm1.4$ & $77.4\pm1.6$ & $52.4\pm0.0$ & $78.2\pm1.8$ & $71.9\pm0.00$ & $76.5\pm1.7$ & $76.7\pm0.0$ \\
  \textbf{Overall Score} $\uparrow$
    & $61.4\pm1.2$ & $67.2\pm1.3$ & $72.8\pm1.5$ & $77.5\pm1.7$ & $68.1\pm0.0$ & $54.2\pm1.4$ & $81.7\pm0.00$ & $69.1\pm1.3$ & \textcolor{red}{\textbf{$86.0\pm0.0$}} \\

  \bottomrule
\end{tabular}
  }
  \label{tab:objects_st2}
\end{table*}

\begin{table*}[t]
  \centering
  \caption{Comparison of unlearning methods for simultaneous unlearning of five artist styles.
  }
  \resizebox{\textwidth}{!}{%
\begin{tabular}{lccccc}
  \toprule
  \multirow{2}{*}{\textbf{Method}}
    & \multicolumn{2}{c}{\textbf{Artist unlearning}}
    & \multicolumn{2}{c}{\textbf{MS COCO-30K}}
    & \multirow{2}{*}{\textbf{Rank $\downarrow$}} \\
  \cmidrule(lr){2-3}\cmidrule(lr){4-5}
    & \textbf{CLIP$_a$ $\downarrow$}
    & \textbf{FSR $\uparrow$}
    & \textbf{FID $\downarrow$}
    & \textbf{CLIP $\uparrow$}
    & {} \\
  \midrule
  FMN
    & $30.27\pm0.16$
    & $52.8\pm4.2$
    & $21.4\pm0.11$
    & $30.82\pm0.09$
    & $4.75$ \\

  ESD
    & $33.62\pm0.08$
    & $39.2\pm5.1$
    & $17.1\pm0.12$
    & $30.52\pm0.10$
    & $6.50$ \\

  UCE
    & $31.89\pm0.0$
    & $44.0\pm0.0$
    & $19.7\pm0.0$
    & $31.19\pm0.0$
    & $5.50$ \\

  AC
    & $33.59\pm0.23$
    & $45.2\pm6.2$
    & $16.6\pm0.21$
    & $31.28\pm0.17$
    & $4.00$ \\

  CP
    & $27.90\pm0.0$
    & $79.6\pm0.0$
    & $18.4\pm0.0$
    & $29.76\pm0.0$
    & $4.50$ \\

  MACE
    & $30.98\pm0.12$
    & $57.4\pm3.5$
    & $15.9\pm0.14$
    & $30.14\pm0.18$
    & $3.75$ \\

  SPM
    & $31.10\pm0.17$
    & $40.0\pm5.9$
    & $17.4\pm0.18$
    & $31.33\pm0.14$
    & $4.50$ \\

  \rowcolor{gray!20}
  \emph{FIA} (Ours)
    & $27.45\pm0.0$
    & $83.4\pm0.0$
    & $16.7\pm0.0$
    & $30.56\pm0.00$
    & \textcolor{red}{\textbf{2.50}}   \\
  \midrule
  SD v1.5
    & 42.10
    & --
    & 14.5
    & 31.34
    & -- \\
  \bottomrule
\end{tabular}

  }
  \label{tab:art_st}
\end{table*}

\begin{figure}[t]
  \centering
  \begin{minipage}{0.95\textwidth}
    \centering
    \small
    \setlength{\tabcolsep}{4pt}
    \renewcommand{\arraystretch}{0.9}
    \captionof{table}{Lists of ImageNet classes used for extended multi-concept unlearning experiments. We refer to the prompt composition of the Imagenette dataset and fix the seed to 0.}
    \label{tab:s}
    \begin{tabularx}{\linewidth}{@{}>{\centering\arraybackslash}m{3cm}
                                    |>{\centering\arraybackslash}X@{}}
        \toprule
        \textbf{Category} & \textbf{Classes} \\
        \midrule
        Animal & German shepherd, Golden retriever, Persian cat, Elephant, Zebra, Horse, Duck \\
        \midrule
        Transportation & Sports car, Minivan, Bicycle, Motorcycle, Bus, Train, Airplane, Ship \\
        \midrule
        Appliance & Keyboard, Computer mouse, Coffee mug, Chair, Sofa, Dining table, Refrigerator, Microwave, Toaster, Television \\
        \midrule
        Daily Items & Bottle, Cell phone, Umbrella, Backpack, Handbag, Suitcase, Sunglasses, Watch \\
        \midrule
        Food \& Drink & Pizza, Hot dog, Banana, Ice cream, Strawberry, Lemon, Pineapple \\
        \bottomrule
    \end{tabularx}
  \end{minipage}

  \vspace{8pt} 

  \begin{minipage}{0.95\textwidth}
    \centering
    \includegraphics[width=\linewidth]{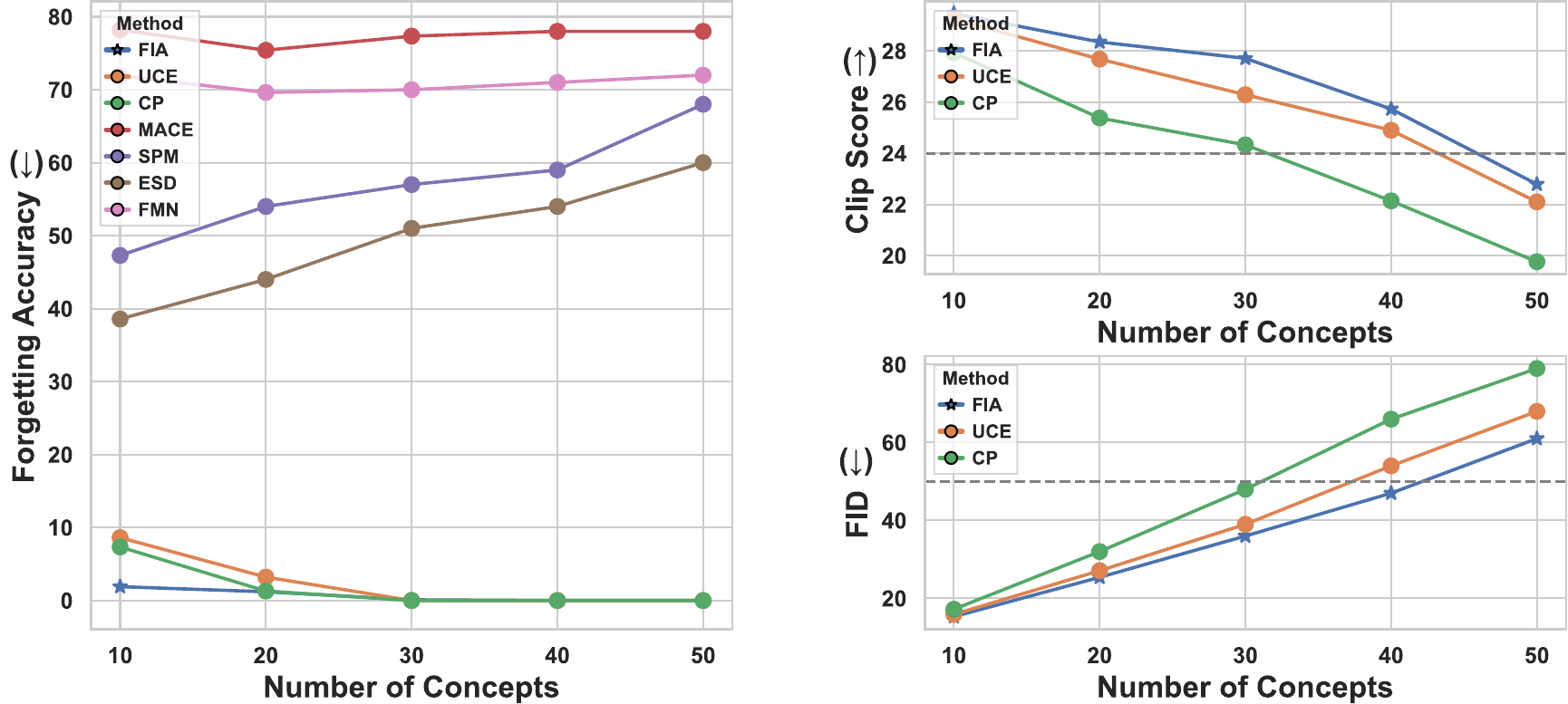}
    \caption{Quantitative results of different methods for unlearning 50 target classes.}
    \label{app:s}
  \end{minipage}
\end{figure}

\clearpage
\newpage
\section{Open Source Code Reference}
\label{app:code}

For fair and reproducible comparison, we benchmark our method against the most relevant state-of-the-art unlearning and concept editing approaches. 
We rely on their official open-source implementations, listed below, and report all baseline results using the recommended configurations unless otherwise specified.

\begin{itemize}
    \item FMN: \url{https://github.com/SHI-Labs/Forget-Me-Not}
    \item SPM: \url{https://github.com/Con6924/SPM}
    \item ESD: \url{https://github.com/rohitgandikota/erasing}
    \item MACE: \url{https://github.com/Shilin-LU/MACE}
    \item UCE: \url{https://github.com/rohitgandikota/unified-concept-editing}
    \item AC: \url{https://github.com/nupurkmr9/concept-ablation}
    \item SalUn: \url{https://github.com/OPTML-Group/Unlearn-Saliency}
    \item CP: \url{https://github.com/ruchikachavhan/concept-prune}
    
\end{itemize}

\end{document}